\documentclass[letterpaper,10pt,journal]{IEEEtran}

\usepackage[numbers]{natbib}
\usepackage{url}					  
\usepackage{graphics,graphicx,color}	
\let\labelindent\relax				
\usepackage{enumitem}	 
\usepackage{gensymb}		
\usepackage[table, usenames, dvipsnames]{xcolor}
\usepackage{geometry}
\usepackage{pifont}
\usepackage{pdflscape}
\usepackage{afterpage}
\usepackage[normalem]{ulem}
\usepackage{soul}

\usepackage{booktabs}
\usepackage{rotating}
\usepackage{csquotes}

\usepackage{hyperref}

\usepackage[hang,flushmargin]{footmisc}

 \usepackage[textwidth=2cm,colorinlistoftodos,textsize=scriptsize]{todonotes}

\newcommand{\JText}[1]{\textcolor{black}{#1}}
\newcommand{\BText}[1]{\textcolor{black}{#1}}
\newcommand{\KText}[1]{\textcolor{black}{#1}}

 \newcommand{\cmark}{\ding{51}}%
 \newcommand{\xmark}{\ding{55}}%

\begin{document}
\newgeometry{margin=14.32mm}
\title{Interactive Perception: Leveraging Action in Perception and Perception in Action}

\author{Jeannette~Bohg*,~\IEEEmembership{Member,~IEEE,}
  Karol~Hausman*,~\IEEEmembership{Student Member,~IEEE,}
  Bharath~Sankaran*,~\IEEEmembership{Student Member,~IEEE,} 
  Oliver~Brock,~\IEEEmembership{Senior Member,~IEEE,}
  Danica~Kragic,~\IEEEmembership{Fellow,~IEEE,}
  Stefan~Schaal,~\IEEEmembership{Fellow,~IEEE,}
  and~Gaurav~Sukhatme,~\IEEEmembership{Fellow,~IEEE}
  \thanks{*-authors contributed equally and are listed alphabetically}
\thanks{K.~Hausman, B.~Sankaran, S.~Schaal and G.~Sukhatme are with the department of Computer Science, University of Southern California, Los Angeles, CA 90089, USA {\tt\small\{khausman, bsankara, sschaal, gsukhatme\}@usc.edu}.}%
\thanks{J.~Bohg and S.~Schaal are with Autonomous Motion Department, Max Planck Institute for Intelligent Systems, T{\"u}bingen, Germany, {\tt\small first.lastname@tuebingen.mpg.de}.}%
\thanks{J.~Bohg is with the Computer Science Department, Stanford University, CA 94305, USA.}
\thanks{D.~Kragic with the Centre for Autonomous Systems, Computational Vision and Active Perception Lab, Royal Institute for Technology (KTH), Stockholm, Sweden {\tt\small dani@kth.de}.}
\thanks{O.~Brock is with Robotics and Biology Laboratory, Technische Universit{\"a}t Berlin, Germany {\tt\small oliver.brock@tu-berlin.de}.}
}

\markboth{IEEE Transactions on Robotics,~Vol.~X, No.~X, Month~20XX}%
{Bohg \MakeLowercase{\textit{et al.}}: Interactive Perception: Leveraging Action in Perception and Perception in Action}

\maketitle
\begin{abstract}
  Recent approaches in robot perception follow the insight that perception is
  facilitated by interaction with the environment. These approaches
  are subsumed under the term Interactive Perception (IP).
  This view of perception
  provides the following benefits. First, interaction with the
  environment creates a rich sensory signal that would otherwise not
  be present. Second, knowledge of the regularity in the combined
  space of sensory data and action parameters facilitates the
  prediction and interpretation of the sensory signal. In this survey, we
  postulate this as a principle for robot perception and collect
  evidence in its support by  
  analyzing and categorizing existing work in this area. We also
  provide an overview of the most important applications of IP. We
  close this survey by discussing remaining 
  open questions. With this survey, we hope to help define the field of
  Interactive Perception and to provide a valuable resource for future
  research. 
\end{abstract}


\IEEEpeerreviewmaketitle


\section{Introduction}
\label{sec:intro}
\IEEEPARstart{T}{here} is compelling evidence that perception in
humans and animals is an active and exploratory
process~\cite{Gibson1979,oReagan2001,Noe2004}.  
Even the most basic categories of biological vision seem to be based
on active visual exploration, rather than on the analysis of static image
content. For example, \citet{Noe2004} argues that the visual category
{\em circle\/} or {\em round\/} cannot be based on the direct
perception of a circle, as (i)~we rarely look at round objects from
directly above,  and (ii)~the projection of a circle onto our retina
is not a circle at all.  Instead, we perceive circles by the way their projection changes \BText{in
  response to eye movements.}

\begin{figure}[ht]
  \begin{center}
    \includegraphics[width=1.0\columnwidth]{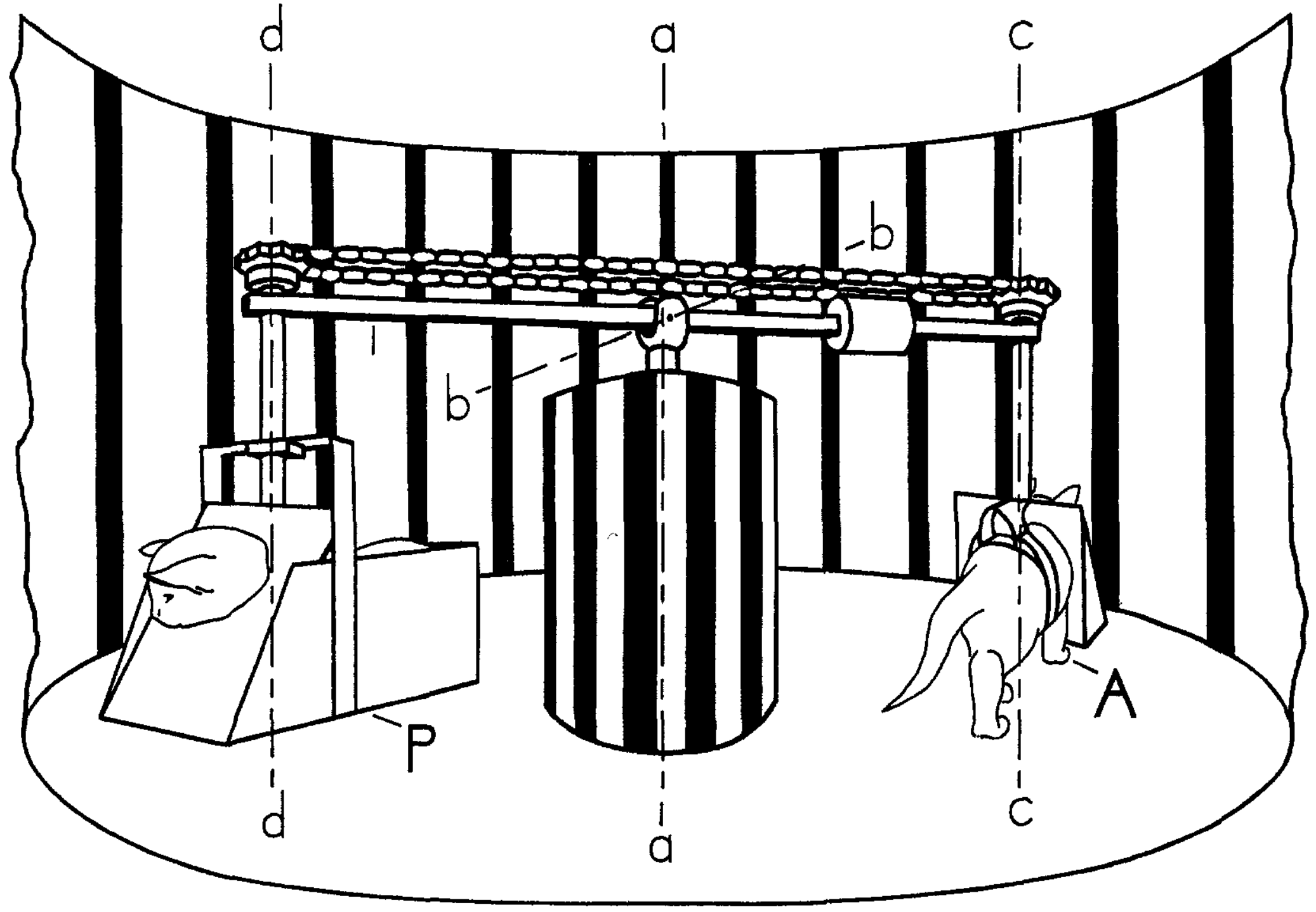}
  \end{center}
  \caption{\BText{A mechanical system where movement of Kitten A is replicated onto Kitten P.
      Both kittens receive the same visual stimuli. Kitten A controls
      the motion, i.e. it is {\em active\/}. Kitten
      P is moved by Kitten A, i.e. it is {\em passive\/}. Only the active kittens developed meaningful
      visually-guided behavior that was tested in separate tasks.
      Figure adapted from~\citet{HeldHein63}.}}
  \label{fig:heldhein}
\end{figure}

\begin{figure}[ht]
  \begin{center}
    \includegraphics[width=1.0\columnwidth]{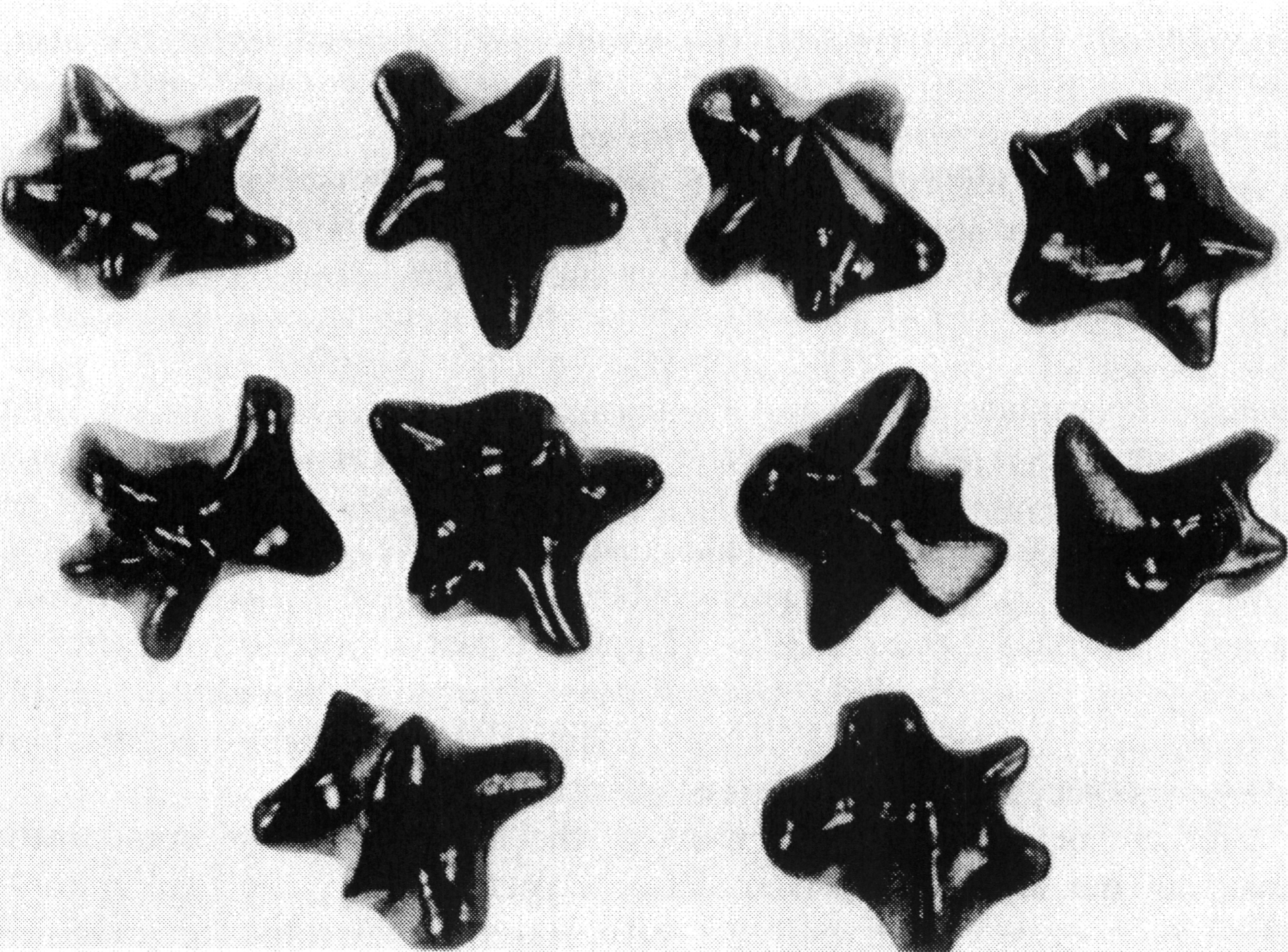}
  \end{center}
  \caption{\BText{Set of irregularly-shaped objects among which subjects had
    to find a reference object. Subjects achieved near perfect
    performance when they could touch and rotate these objects as opposed
    to just looking at them in a static pose. Figure adapted
    from~\citet[p.124]{Gibson1966} with permission.}}  
  \label{fig:jjgibson}
\end{figure}

\BText{\citet{HeldHein63} analyzed the development of visually-guided
  behavior in kittens. They found that this development critically
  depends on the opportunity to learn the relationship between
  self-produced movement and concurrent visual feedback. The
  authors conducted an experiment with kittens that were only exposed
  to daylight when placed in the carousel depicted in
  Fig.~\ref{fig:heldhein}. Through this mechanism, the
  {\em active\/} kittens (A) transferred their own,
  deliberate motion to the {\em passive\/} kittens (P)
  that were sitting in a basket. Although, both types of kittens received the
  same visual stimuli, only the active kittens showed meaningful
  visually-guided behavior in test situations.} 

\citet{Gibson1966} showed that
physical interaction further augments perceptual processing beyond
what can be achieved by deliberate pose changes.
In the specific experiment, human subjects had to find a reference
object among a set of irregularly-shaped, three-dimensional objects
(see Fig.~\ref{fig:jjgibson}). They achieved an average accuracy of
49\% if these objects were shown in a single image.  
This accuracy increased to 72\% when subjects viewed rotating versions
of the objects. They achieved nearly perfect performance (99\%)
when touching and rotating the objects in their hands.  

\BText{ These three examples illustrate that biological perception and
  perceptually-guided behavior intrinsically rely on active
  exploration and knowledge of the relation between action
  and sensory response. This contradicts our
  introspection, as we just seem to passively {\em see\/}. In reality, 
  visual perception is similar to haptic exploration.  ``{\em Vision
    is touch-like\/}''~\citep[p.73]{Noe2004} in that, perceptual
  content is not given to the observer all at once but only through
  skillful, active {\em looking.\/}}

\BText{This stands in stark contrast to how perception problems are commonly
  framed in Machine Vision. Often, the aim is to
  semantically annotate a single image while relying on a minimum set
  of assumptions or prior knowledge. These requirements render the
  considered perception problems under-constrained and thereby make them
  very hard to solve.}

  \BText{The most successful approaches learn models from data 
  sets that contain hundreds of thousands of semantically annotated
  \textit{static} images, such as Pascal VOC~\citep{Everingham:2010},
  ImageNet~\citep{ILSVRCarxiv14} or Microsoft
  COCO~\citep{LinMBHPRDZ14}. Recently, Deep Learning based
  approaches led to substantial progress by being able to leverage these
  large amounts of training data. In these methods, data points
  provide the most important source of constraints to find a suitable solution to the
  considered perception problem. The success of these methods over
  more traditional approaches suggests that previously considered
  assumptions and prior knowledge did not correctly or 
  sufficiently constrain the solution space.} 

\BText{Different from disembodied Computer Vision algorithms, robots
  are embodied agents that can move within the environment and
  physically interact with it. Similar to biological systems, this
  creates rich and more informative sensory signals that are
  concurrent with the actions and would otherwise not be
  present. There is a regular relationship between actions and their
  sensory response. This regularity provides the additional constraints that
  simplify the prediction and interpretation of these high-dimensional signals.
  Therefore, robots should exploit any knowledge of this regularity. Such an
  integrated approach to perception and action may reduce the
  requirement of large amounts of data and thereby provide a viable
  alternative to the current data-intensive approaches towards machine
  perception.} 


\section{Interactive Perception}
\BText{Recent approaches in robot perception are subsumed by the term {\em
    Interactive Perception\/} (IP). They exploit any kind of forceful
  interaction with the environment to simplify and enhance
  perception. Thereby, they
  enable robust perceptually-guided manipulation behaviors. IP has two
  benefits. First, physical interaction creates a novel
  sensory signal that would otherwise not be present. Second, by
  exploiting knowledge of the regularity in the combined space of
  sensory data and action parameters, the prediction and
  interpretation of this novel 
  signal becomes simpler and more robust. In this section, we will
  define what we mean by forceful interaction. Furthermore, we explain
  the two postulated benefits of IP in more detail.}
 
\subsection{Forceful Interactions}
Any action that exerts a potentially \BText{time-varying} force upon
the environment is a forceful interaction. A common way of creating such an
interaction is through physical contact that may \BText{be established
  for the purpose of moving the agent (e.g. in legged or wheeled
  locomotion), for changing the environment (e.g. to open a door or
  pushing objects on a table out of the way) or for exploring
  environment properties while leaving it unchanged (e.g. by sliding
  along a surface to determine its material).}  
It may also be a contact-free interaction that is due to gravitational
or magnetic forces \BText{or even lift. An interaction may only be
  locally applied to the scene (e.g. through pushing or pulling a
  specific object) or it may affect the scene globally (e.g. shaking a
  tray with objects standing on it).}
This interaction can be performed either by the agent itself or by any
other entity, e.g. a teacher to be imitated or someone who
demonstrates an interaction through kinesthetic teaching.

\BText{In this survey we are interested in
approaches that go beyond the mere observation of the environment
towards approaches that enable its} {\em \BText{Perceptive
      Manipulation\/}}\footnote{\BText{We consider {\em Perceptive
        Manipulation\/} to be the equivalent term to Interactive
      Perception. This emphasizes the blurred boundary which is
      traditionally drawn between manipulation and perception.}}.
  Therefore, we focus on physical interactions for the purpose of
changing the environment or for exploring environment properties while
leaving it unchanged. We are not concerned with interactions for
locomotion and environment mapping. 

\subsection{Benefits of Interactive Perception}
\label{sec:conditions}
\subsubsection*{Create Novel Signals (CNS)}
\BText{Forceful interactions create novel, rich sensory signals that would 
otherwise not be present. These signals are beneficial for estimating
the quantities that are relevant to manipulation problems such as haptic, audio and
visual data correlated over time. Relevant quantities include object
weight, surface material or rigidity.}

\subsubsection*{Action Perception Regularity (APR)}
Forceful interactions reveal regularities in the combined space ($S \times A
  \times t$) of sensor information ($S$) and action parameters ($A$)
  over time ($t$). \BText{This regularity is constituted by the
      repeatable, multi-modal sensory data that is created when executing the
      same action in the same environment. Not considering the space of
      actions amounts to marginalizing over them. The corresponding sensory signals
      would then possess a significantly higher degree of variation compared to
      the case where the regularity in $S \times A \times t$ is taken
      into account. Therefore despite $S \times A \times t$ being much higher dimensional, the signal represented in this space has more structure.} 
  \paragraph*{Using the Regularity} \BText{Knowing this regularity
    corresponds to understanding the causal relationship between
    action and sensory response given specific environment properties. 
    Thereby, it allows to (i) predict the sensory signal given
    knowledge about the action and environment properties,  
    (ii) update the knowledge about some latent properties of the
    environment by comparing the prediction to the observation and  
    (iii) infer the action that has been applied to generate the
    observed sensory signal given some environment properties. These
    capabilities simplify perception but also enable optimal action
    selection.} 
  \paragraph*{Learning the Regularity} \BText{Learning these
    regularities corresponds to identifying the 
    causal relationship between action and sensory response. This
    requires information about the action that produced an observed
    sensory effect. If the robot autonomously interacts with the
    environment, this information is automatically
    available. Information about the action can also be provided by
    a human demonstrator.}
\section{\BText{Historical Perspective}}\label{sec:active_percep}
\BText{In robotics, the research field of {\em Active Perception\/} (AP) pioneered the
  insight that perception is active and exploratory.  
  In this section, we relate Interactive to Active
  Perception. Additionally, we discuss the relation of IP to other
  perception approaches that neglect either the sensory or
  action space in $S \times A \times t$. Figure~\ref{tab:perception}
  summarizes the section.}  
  
\begin{figure}
  \centering
  \begin{tabular}{ p{5.2cm}m{0.3cm}m{0.3cm}m{0.3cm}m{0.3cm} }
    \toprule
    & $S$ & \multicolumn{2}{c}{$A$} & $t$ \\
    \cmidrule(r){3-4}
    &  & $F$ & $\neg F$  &   \\
    \midrule
    Sensorless Manipulation & - & \cmark & - & \cmark \\\hline
    Perception of Images & \cmark & - & - & - \\\hline
    Perception of Video & \cmark & - & - & \cmark \\\hline
    Active Perception & \cmark & - & \cmark & \cmark \\\hline
    Active Haptic Perception, \newline Interactive Perception  & \cmark & \cmark & - & \cmark \\
    \bottomrule
  \end{tabular}
  \caption{Summary of how Interactive Perception relates to other
    perception approaches regarding $S \times A \times t$. $F$ stands
    for forceful interaction and $\neg F$ for actions that only manipulate
    the parameters of the sensory apparatus and not the
    environment. 
    }
  \label{tab:perception}
\end{figure}

\subsection{\BText{Sensorless Manipulation}}
\BText{This approach to perception does not require any sensing. It
  aims at finding} a sequence of actions that brings the
system of interest from an unknown into a defined state. Therefore,
after performing these actions, the \BText{system} state
\BText{can be considered as perceived.} 
This kind of sensorless manipulation was demonstrated first by
  \citet{Erdmann:88} who used it for orienting a planar part that is
  randomly dropped onto a tray.
The goal of the proposed algorithm is to generate a sequence of tray
tilting actions that uniquely moves the part into a goal orientation
without receiving sensor feedback or knowing the initial state. It
uses a simple mechanical model of sliding and
information on how events like collisions with walls reduce the number
of possible part orientations. More recently, \citet{DogarRSS12}
extend this line of thought to grasping.  
The authors plan for the best push-grasp such that the object of
interest has a high probability of moving into the gripper while other
objects are pushed away. The plan is then executed open-loop without
taking feedback of the actual response of the environment into
account.  

\BText{We argue that IP critically depends on representing a signal in
  the combined space of sensory information and action parameters over
  time. Sensorless manipulation is similar in that it also requires a
  model of how actions funnel the uncertainty about the system state
  into a smaller region in state space. However, different from the approaches in this
  survey, it does not require sensory feedback as it assumes that
  the uncertainty can be reduced to the required amount only through the
  actions. For complex dynamical systems, this may not always be the
  case or a sufficiently expressive forward model may not be available.} 

\subsection{\BText{Perception of Visual Data}}
\BText{The vast research area of Computer Vision focuses on
  interpreting static
  images, video or other visual data. The majority of approaches
  completely neglect the active and exploratory nature of human and
  robot perception. Nevertheless, there are examples in the
  Computer Vision literature that show how exploiting the regularity in
  $S\times A \times t$ simplifies perception problems. The
  first example aims at human activity recognition in
  video. It is somewhat obvious that this task becomes easier when observing the activity
  over a certain course of time. Less obvious is the result by
  \citet{Kjellstrom:CVIU:2010} who showed that classifying objects is
  easier if they are observed while being used by a person. More
  recently, \citet{Cai-RSS-16} support these results. They show that
  recognizing manipulation actions in single images is much easier
  when modeling the associated grasp and object type in a unified model.} 

\BText{Another example considers the problem of image
  restoration. \citet{Xue:2015} exploit whole image sequences to
  separate obstructing foreground like fences or window reflections
  from the main subject of the images, i.e. the background. This would
  be a very hard problem if only a single image were given or without the
  prior knowledge of the relation between optical flow and depth.}

\BText{\citet{Aloimonos88} show how challenging vision problems, such as
  shape from shading or structure from motion, are
  easier to solve with an active than a passive observer. Given known
  camera motion and associated images, the
  particular problem can be formulated such that it has a unique
  solution and is linear. The case of the passive observer usually
  requires additional assumptions or regularization and sometimes
  non-linear optimization.}

\subsection{\BText{Active Perception}}
\BText{In 1988, \citet{BajcsyAT88} introduced AP as the problem of
  intelligent control strategies applied to the data acquisition
  process. \citet{ballard1991animate} and \citet{Aloimonos88} further
  analyzed this concept for the particular modality of vision. In this
  context, researchers
  developed artificial vision systems with many degrees of
  freedom~\citep{pahlavan1993active,YorickB98,ArmarHead} and models of
  visual attention~\citep{Itti_Koch01nrn,Tsotsos1995507} that these
  active vision systems could use for guiding their gaze.}

\BText{Recently, \citet{BajcsyAT16} revisited AP giving an excellent
  historical perspective on 
  the field and a new, broader definition of an active perceiver based
  on decades of  research:}
\begin{displayquote}
{\em \BText{``An agent is an active perceiver if it knows why it wishes to sense, and then chooses what to perceive, and determines how, when and where to achieve that perception.''}\/}
\end{displayquote}

\BText{The authors identify the {\em why\/} as the
  central and distinguishing component to a passive observer. It
  requires the agent to reason about so called {\em
  Expectation-Action\/} tuples to
  select the next best action. The expected result of the action can
  be confirmed by its execution. Expectation-Action tuples capture the
  predictive power of the regularity in $S\times A\times t$ to enable
  optimal action selection.}

  \subsubsection{Relation to Interactive Perception}
  \BText{The new definition of AP is not only restricted to vision. However, the
  majority of approaches gathered 
  under the term of active perception consider vision as the sole
  modality and the manipulation of extrinsic or intrinsic camera
  parameters as possible actions. This is also reflected by the choice
  of examples in~\citep{BajcsyAT16}. 
  The focus on the visual sense has several implications for Active
  Perception in relation to Interactive
  Perception. First, an
  active perceiver with the ability to move creates a richer and more informative visual
  signal (e.g. from multiple viewpoints or when zooming) that would
  otherwise not be present.  However,  
  this may not provide all relevant information, especially not those
  required for manipulation problems. \citet{Metta:HapticRepr:04}
  emphasize that only through physical interaction, a robot can access object
  properties that otherwise would not be available (like weight,
  roughness or softness).}

\BText{Second, as shown in \citet{Aloimonos88}, we have very good
  understanding of multi-view and perspective geometry that can be
  leveraged to formulate a vision problem in such a way that its
  solution is simple and tractable. However, when it comes to
  predicting the effect of physical interaction that does not only
  change the viewpoint of the agent on the environment, but the
  environment itself, we are yet to develop rich, expressive and
  tractable models.}  

\BText{Lastly, AP mainly focuses on simplifying challenging
  perception problems. However, a robot should also be able to manipulate the
  environment in a goal-directed
  manner. \citet[p.167]{Sandini_VisionDuringAction:93} formulate this
  as a difference in how visual information is used: in AP it is
  mainly devoted to exploration of the environment whereas in IP it is
  also used to monitor the execution of motor
  actions.} 

\subsubsection{Early Examples of Interactive Perception}
\BText{There are a number of early approaches within the area of
  Active Perception that exploit forceful interaction with the
  environment and are therefore early examples for IP
  approaches. \citet{Tsikos88,TsikosRAM91} propose to use a robot arm
  to make the scene simpler for the vision system through actions like
  pick, push and shake. The specific scenario is the separation of
  random heaps of objects into sets of similar
  shapes. \citet{Bajcsy:89,Sinha:90} propose the {\em Looker and
    Feeler\/}  system that allows to perform material recognition of
  potential footholds for legged locomotion.  
The authors hand-design specific exploration procedures of which the robot
observes the outcome (visually or haptically) to determine material
attributes. \citet{Salganicoff:91} show how the mapping between
observed attributes, actions and rewards can be learned from training
data gathered during real executions of a task.  
\citet[Section 3]{Sandini_VisionDuringAction:93} propose to use
optical flow analysis of the object motion while it is being pushed.  
The authors show that through this analysis, they can retrieve both
geometrical and physical object properties which can then be used to
adapt the action.}  

\subsection{\BText{Active Haptic Perception}}

Haptic exploration of the environment relies on haptic sensing that
requires contact with the environment. Interpretation of a sequence of
such observations is part of IP as it requires a forceful and
time-varying interaction. The interpretation of an isolated haptic
{\em frame\/} \BText{without temporal information is similar to
  approaches in Computer Vision such as semantic scene understanding from
  static images~\citep{TactileSensing:Survey:89}.} 

 \BText{Early approaches that use touch in an active manner are
   applied to problems such as reconstructing shape from
   touch~\citep{Bajcsy:84:ShapeFromTouch}, recognizing objects through
   tracing their surface~\citep{BajcsyGoldberg:84} or exploring
   texture and material
   properties~\citep{Bajcsy:84:ShapeFromTouch}. The complementary
   nature of vision and touch has been explored by~\citet{Allen:85} in
   reconstructing 3D object shape. 
A more complete review of these early approaches towards active haptic
perception is contained
in~\citep{TactileSensing:Survey:89,BajcsyAT16}.} 

\BText{More recent examples include \citep{Metta:HapticRepr:04}
  to learn haptic object representations, \citep{Petrovskaya:2011,
    HerbertICRA2013,JavdaniICRA2013} for object detection and pose
  estimation, \citep{DragievICRA2011, KimCAS2014,bohg:iros10} for
  reconstructing the shape of objects or the environment as well as
  \citep{FishelLoeb:2012, LoebFishel:2014,ChuICRA13} for texture
  classification or description. The most apparent difference of these 
  recent approaches to earlier work lies in the use of machine
  learning techniques to either automatically find suitable
  exploration strategies, to learn suitable feature representations or
  to better estimate different quantities.}

\BText{In general, active haptic perception requires deliberate
  contact interaction but the majority of the cases do not aim at
  changing the environment. Instead, for simplification, objects or
  the environment are often assumed to be rigid and static during
  contact.}

\begin{figure*}
\begin{tikzpicture}[scale=0.9]
 
 \draw[red,thick](7,0) -- (15.7,-1.5);
  \node[color=red] at (11.7,1.4){{\large \bf Object Segmentation }};
  \node[color=red!80!green] at (10.7,-0.7){\footnotesize\citet{KatzICRA2011}};
  \node[color=red] at (13.0,0.5){\footnotesize\citet{BergstromICVS11}};
  \node[color=red] at (13.7,0.2){\footnotesize\citet{BerschRSS-W12}};
  \node[color=red] at (9.9,0.2){\footnotesize\citet{SchiebenerICRA14}};
  \node[color=red] at (13.5,-0.2){\footnotesize\citet{HoofIROS12,hoofHUMANOIDS13}};
  \node[color=red] at (14.5,-0.7){\footnotesize\citet{xu_siga15}};  
  \node[color=red] at (13.5,2.2){\footnotesize\citet{GuptaICRA12}};
  \node[color=red] at (13.9,0.9){\footnotesize\citet{KarolICRA13}};  
  \node[color=red] at (13.0,2.7){\footnotesize\citet{KuzmicHUMANOIDS10}};
  \node[color=red] at (13.0,3.2){\footnotesize\citet{FitzpatrickIROS12}};
  \node[color=red] at (14.0,3.7){\footnotesize\citet{MettaNN03}};
  \node[color=red] at (14.0,4.2){\footnotesize\citet{KenneyICRA09}};%
  \node[color=red] at (14.2,4.7){\footnotesize\citet{ChangICRA12}};    
  \node[color=red!60!violet] at (9.8,1.9){\footnotesize\citet{Schiebener2011}};
  \node[color=red!60!violet] at (8.2,1.0){\footnotesize\citet{UdeIJHR08}};
 
 \draw[violet,thick](7,0) -- (15.0,6.5);
 \node[color=violet] at (9.2,3.9){{\large \bf  Object}};
 \node[color=violet] at (9.0,3.4){{\large \bf  Recognition }};
 \node[color=violet] at (8.8,2.8){\footnotesize\citet{SinapovICRA14,SinapovRAS14}}; 
 \node[color=violet] at (11.0,5.2){\footnotesize\citet{hausman14iros-ws-b}};
 \node[color=violet] at (11.0,5.7){\footnotesize\citet{SinapovICRA13}};
 \node[color=violet] at (10.8,4.4){\footnotesize\citet{Kjellstrom:CVIU:2010}};
 
 \draw[blue!60!black!80!violet,thick](7,0) -- (7.6,6.2);
   \node[color=blue!60!black!80!violet] at (3.2,5.7){{\large \bf Manipulation}};
   \node[color=blue!60!black!80!violet] at (3.40,5.2){{\large \bf Skill Learning}};
   \node[color=blue!60!black!80!violet] at (5.2,4.0){\footnotesize\citet{Cusumano-TownerSMOA11}};
   \node[color=blue!60!black!80!violet] at (4.2,4.5){\footnotesize\citet{LeeICRA15}};   
   \node[color=blue!60!black!80!violet] at (5.2,3.5){\footnotesize\citet{pastor2011online}};
   \node[color=blue!60!black!80!violet] at (5.2,3.0){\footnotesize\citet{Kappler-RSS-15}};
   \node[color=blue!60!black!80!violet] at (5.1,2.0){\footnotesize\citet{LevineICRA15}};
   \node[color=blue!60!black!80!violet] at (5.2,2.5){\footnotesize\citet{HanIROS15}};

  \draw[blue!60!black!80!green,thick](7,0) -- (-2.5,5.7);
  \node[color=blue!60!black!80!green] at (-2.2,3.0){{\large \bf State Representation}};
  \node[color=blue!60!black!80!green] at (-2.2,2.5){{\large \bf Learning}};  
  \node[color=blue!60!black] at (-2.5,1.9){\footnotesize\citet{finn2016unsupervised}};
  \node[color=blue!60!black] at (0.7,1.2){\footnotesize\citet{AgrawalNAML16}};
  \node[color=blue!60!black!80!green] at (0.8,1.9){\footnotesize\citet{Jonschkowski-14-AR}};
  \node[color=blue!60!black!80!green] at (-2.8,4.5){\footnotesize\citet{pinto2016curious}};  
  \node[color=blue!60!violet!80!green] at (-0.2,4.0){\footnotesize\citet{AssaelWSD15}}; 
  \node[color=blue!60!violet!80!green] at (0.6,3.5){\footnotesize\citet{LevineFDA15}};
 
  \draw[blue,thick](7,0) -- (-4.5,2);
  \node[color=blue] at (-2.2,0.5){{\large \bf  Object Dynamics }};
  \node[color=blue] at (-2.2,0.0){{\large \bf  Learning}};
  \node[color=blue] at (1.25,0.20){\footnotesize\citet{Atkeson01091986}};
  \node[color=black!60!cyan!80!blue] at (2.1,-0.3){\footnotesize\citet{ZhangICRA12}};%
  \node[color=black!60!cyan!80!blue] at (-0.7,-0.5){\footnotesize\citet{NIPS2015_Galileo}};
  
 \draw[cyan,thick](7,0)-- (-4.5,-0.7);
    \node[color=cyan] at (-2.7,-2.2){{\large \bf Pose Estimation }};
    \node[color=cyan] at (-0.2,-1.1){\footnotesize\citet{KovalIROS13}};
    \node[color=cyan] at (-3.2,-1.5){\footnotesize\citet{Mason1990}};
    \node[color=cyan] at (-0.8,-1.85){\footnotesize\citet{JavdaniICRA2013}};    
    \node[color=black!30!green!30!cyan] at (3.0,-1.41){\footnotesize\citet{KovalRSS14}};    
    \node[color=black!30!green!30!cyan] at (-2.0,-3.2){\footnotesize\citet{LPKTLPICRA12}}; %
 
  \draw[cyan!30!green,thick](7,0) -- (-3.5,-3.5);
   \node[color=cyan!30!green] at (1.5,-2.7){{\large \bf Grasp Planning}};
   \node[color=cyan!30!green] at (2.7,-2.0){\footnotesize\citet{DogarICRA13}};
   \node[color=cyan!30!green] at (-0.7,-3.7){\footnotesize\citet{PlattISRR11}};
   \node[color=cyan!30!green] at (-0.5,-4.3){\footnotesize\citet{HsiaoICRA2009}}; %
   \node[color=cyan!30!green] at (-3.0,-4.3){\footnotesize\citet{kroemer2010combining}};
   \node[color=cyan!30!green] at (-1.0,-4.7){\footnotesize\citet{Boularias_AAAI_2015}};
   \node[color=green!30!black!60!blue] at (3.2,-3.5){\footnotesize\citet{DragievICRA2013}};

  \draw[blue!60!black!60!green,thick](7,0) -- (0.2,-5.5);
   \node[color=blue!60!black!60!green] at (3.3,-5.0){{\large \bf Multimodal Object}};
   \node[color=blue!60!black!60!green] at (3.0,-5.5){{\large \bf Model Learning}};
   \node[color=blue!60!black!60!green] at (4.0,-3.9){\footnotesize\citet{DragievICRA2011}};
   \node[color=blue!60!black!60!green] at (6.9,-4.1){\footnotesize\citet{SinapovICRA14, SinapovRAS14}};
   \node[color=blue!60!black!60!green] at (3.0,-6.0){\footnotesize\citet{Kra11Aut}};
   \node[color=blue!60!black!60!green] at (7.2,-4.5){\footnotesize\citet{SinapovICRA13}};
   \node[color=blue!60!black!60!green] at (5.0,-2.7){\footnotesize\citet{IlonenBK14}};
   \node[color=blue!60!black!60!green] at (4.0,-4.3){\footnotesize\citet{bohg:iros10}};
   \node[color=blue!60!black!60!green] at (4.8,-3.1){\footnotesize\citet{BjorkmanBHK13}}; %

  \draw[black!60!green,thick](7,0) -- (5.5,-6.4);
  \node[color=black!60!green] at (8.1,-5.0){{\large \bf Haptic Property}};
  \node[color=black!60!green] at (8.1,-5.5){{\large \bf Estimation}};
  \node[color=black!60!green] at (7.7,-3.6){\footnotesize\citet{CulbertsonUK14}};
  \node[color=black!60!green] at (8.3,-6.0){\footnotesize\citet{RomanoTRO12}};
  \node[color=black!60!green] at (7.6,-3.2){\footnotesize\citet{KroemerTRO11}};%
  \node[color=black!60!green] at (7.6,-2.8){\footnotesize\citet{ChuICRA13}};
  
 \draw[blue!60!green,thick](7,0) -- (10.7,-6.0);
    \node[color=blue!60!green] at (12.2, -4.5){{\large \bf  Articulation Model}};
    \node[color=blue!60!green] at (12.2, -5.0){{\large \bf  Estimation}};
    \node[color=blue!60!green] at (11.0,-4.1){\footnotesize\citet{jain2010pulling}};
    \node[color=blue!60!green] at (11.0,-3.7){\footnotesize\citet{sturm2011probabilistic}};        
    \node[color=blue!60!green] at (11.2,-3.2){\footnotesize\citet{martin_iros_ip_2014}};
    \node[color=blue!60!green] at (13.0,-2.8){\footnotesize\citet{brock2009learning}};
    \node[color=blue!60!green] at (10.7,-2.4){\footnotesize\citet{hausman15icra}};
    \node[color=blue!60!green] at (14.2,-2.0){\footnotesize\citet{Pillai-RSS-14}};    
    \node[color=blue!60!green] at (11.7,-2.0){\footnotesize\citet{BarraganICRA14}};
    \node[color=blue!60!green] at (11.7, -1.6){\footnotesize\citet{otte2014entropy}};
    \node[color=blue!60!green] at (9.4,-1.2){\footnotesize\citet{Karayiannidis2013}};
    \node[color=blue!60!green] at (9.3,-1.5){\footnotesize\citet{KatzISER2010}};

 \end{tikzpicture}

\centering
\caption{Paper categorization based on application areas. Papers that address multiple application areas lie on the boundary between those applications, e.g. \citet{LPKTLPICRA12}, \citet{Schiebener2011}} 
\label{fig:categories}
\end{figure*}
\section{Applications of Interactive Perception}
\label{sec:goals}
\BText{Interactive perception methods may be applied to achieve an estimation or a
manipulation goal. Currently, the vast majority of IP approaches
estimate some quantity of interest through forceful interaction.
Other IP approaches pursue either a grasping or manipulation
goal. This means that they aim to manipulate the environment to
bring it into a desired state. Usually this includes the estimation
of quantities that are relevant to the manipulation task.}

\begin{figure}[h]
  \begin{center}
    \includegraphics[width=1.0\columnwidth]{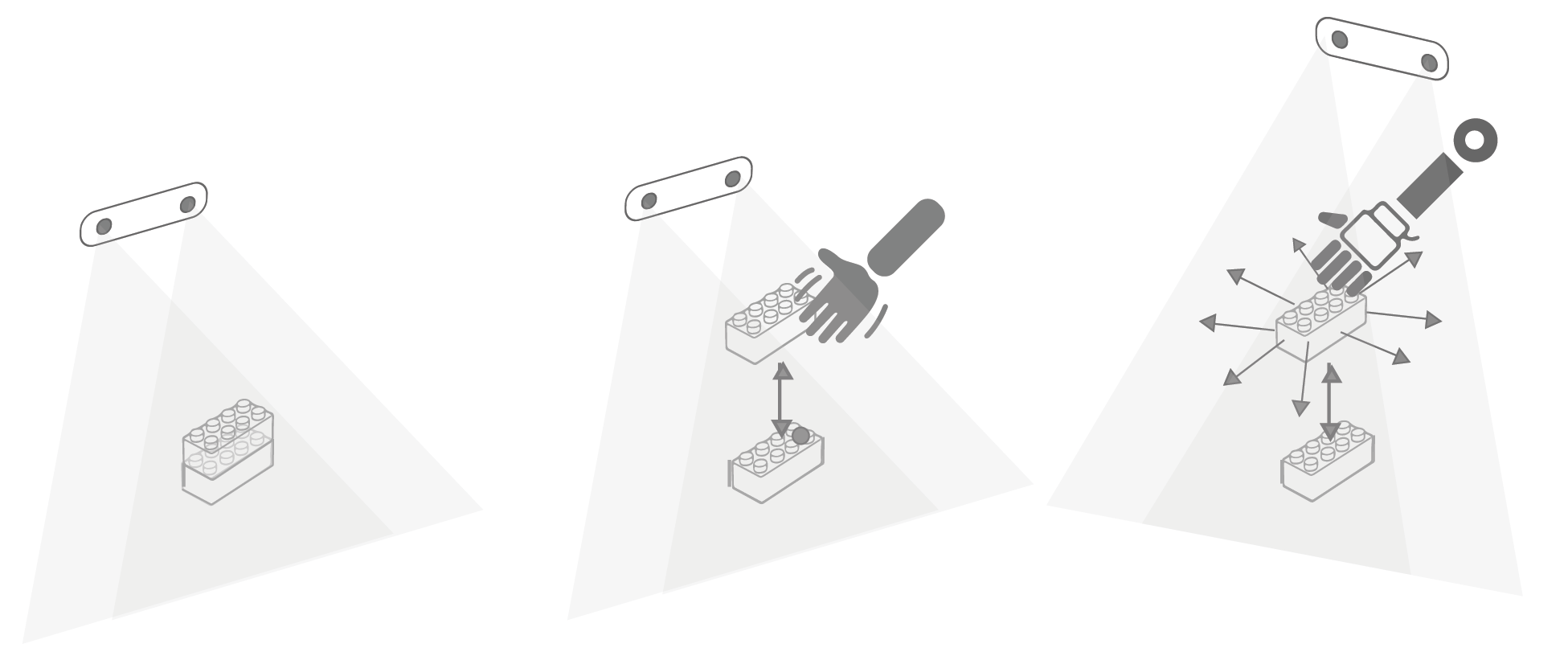}
  \end{center}
  \caption{\BText{Three situations in which a robot (indicated by the
      stereo camera and viewing cones) tries to estimate
      the articulation model of two Lego blocks on a table. The
      situations differ in the amount of information the robot has
      access to. [Left] The robot can only change the viewpoint to
      obtain more information. [Center] The robot can observe a rich sensory
      signal caused by a person lifting the top Lego block. [Right] The
      robot can interact with the scene and observe the resulting sensory
      signal. Therefore, it has more information about the {\em
        specific\/} interaction. Only in the rightmost situation, the
      articulation model can be reliably identified.}}  
  \label{fig:lego}
\end{figure}

\begin{figure}[h]
  \begin{center}
    \includegraphics[width=1.0\columnwidth]{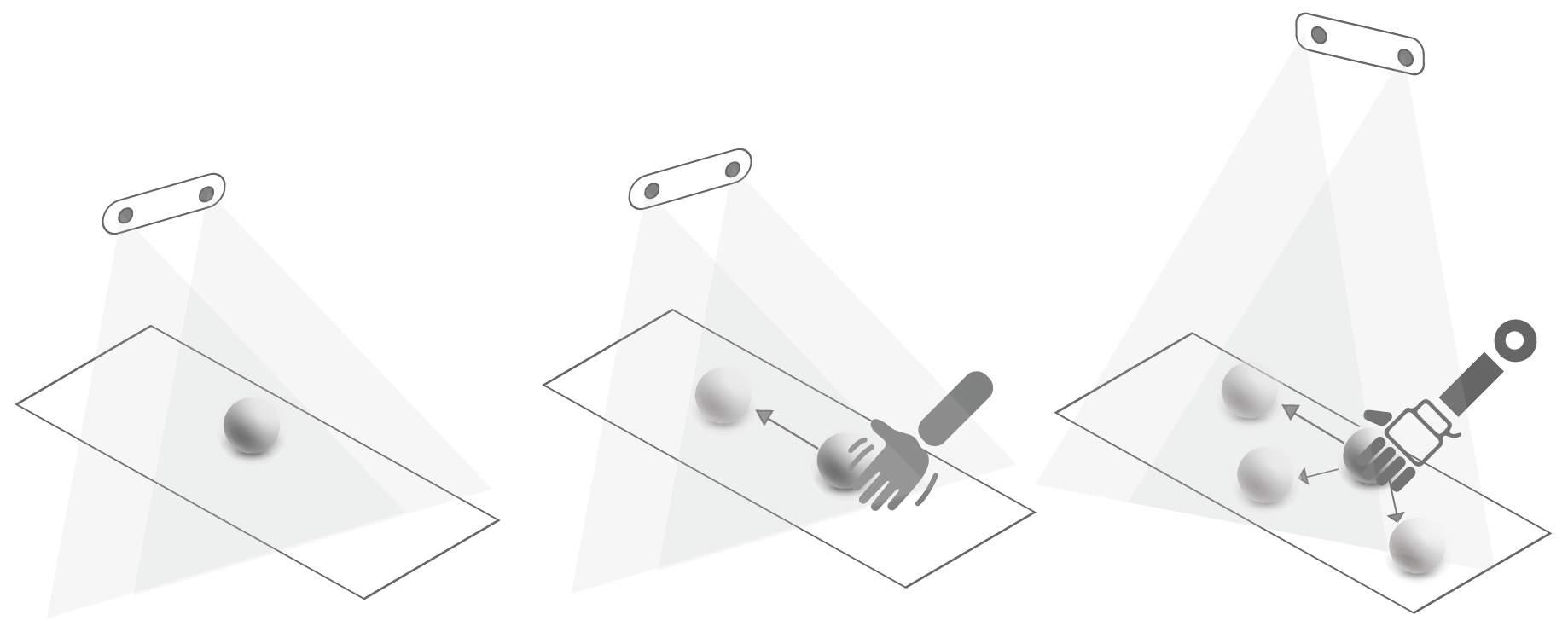}
  \end{center}
  \caption{\BText{Three situations in which a robot (indicated by the
      stereo camera and viewing cones) tries to estimate the weight of
      a sphere. The situations differ in the amount of information the
      robot has access to. [Left] The robot can only change the viewpoint to
      obtain more information. [Center] The robot can observe a rich sensory
      signal caused by a person pushing the sphere. [Right] The
      robot can push the sphere itself and observe the resulting sensory
      signal, i.e. where the sphere comes to rest. In the last
      situation, it has more information about the {\em 
        specific\/} push force. Only in the rightmost situation, the
      weight of the sphere can be reliably unidentified.}}   
  \label{fig:ball}
\end{figure}

\BText{Existing IP approaches can be broadly grouped into ten major
  application areas as visualized in Figure~\ref{fig:categories}. In
  this section, we briefly describe each of these areas. For the first
  three applications (Object Segmentation, Articulation Model
  Estimation and Object Dynamics Learning), we use a couple of simple
  examples (Figures~\ref{fig:lego} and~\ref{fig:ball}) to allow the
  reader to better appreciate the benefits of IP and understand its
  distinction to Active Perception.}

\subsection{Object Segmentation}
\BText{Object segmentation is a difficult problem and, in the area of
Computer Vision, it is often performed on single
images \cite{pami_albaraez_segmentation,NIPS2012_Xiaofeng, LongSD15}.
To illustrate the challenges, consider the simple example scenario depicted in
Figure~\ref{fig:lego}. Two Lego blocks are firmly attached to the
table. The robot is supposed to 
estimate the number of objects on the table. When the robot is a
passive observer of the scene as in Fig.~\ref{fig:lego} [Left], it
would be very challenging to estimate the correct number of Lego
blocks on the table without incorporating a lot of prior
knowledge. The situation does not improve in this static 
scenario even with more sensory data from different viewpoints or
after zooming in.}

\BText{When the robot observes another agent interacting with the
  scene as shown in Fig.~\ref{fig:lego} [Center], it will be able to
  segment the Lego blocks and correctly estimate the number of objects
  in the scene. This is an example of how forceful interactions can
  create rich sensory signals that would otherwise not be present
  (CNS). The new evidence in form of motion cues simplifies the 
  problem of object segmentation.} 

\BText{The ability to interact with the scene allows a robot to also
autonomously generate more informative sensory information as
visualized in Fig.~\ref{fig:lego} [Right]. Reasoning about the
regularity in $S\times A \times t$ may}
lead to even better segmentation since \KText{the robot can select
  actions that are particularly well suited for reducing the
  segmentation uncertainty (APR).}

For these reasons, object segmentation has become a very popular topic
in Interactive Perception. For example, \citet{FitzpatrickIROS12, MettaNN03} are
able to segment the robot's arm and the objects that were moved in a
scene. \citet{GuptaICRA12, ChangICRA12} use predefined actions to
segment objects in cluttered environments. \citet{hoofHUMANOIDS13} can
probabilistically reason about optimal actions to segment a scene.

\subsection{Articulation Model Estimation}
\BText{Another problem that is simplified through Interactive Perception is
the estimation of object articulation mechanisms. The robot has to
determine whether the relative movement  
of two objects is constrained or not. Furthermore, it has to
understand whether this potential constraint is due to a prismatic or
revolute articulation mechanism and what the pose of the joint axis
is. Fig.~\ref{fig:lego} [Left] visualizes an example situation in 
which the robot has to estimate the potential articulation mechanism
between two Lego blocks given only visual observations of a static scene. This is almost
impossible to estimate from single images 
without including a lot of prior semantic 
knowledge. It is also worth noting that this situation is not improved if gathering
more information from multiple viewpoints of this otherwise static
scene.}

\BText{In Fig.~\ref{fig:lego} [Center], the robot observes an agent lifting
the top-most lego block. This is another example 
of how forceful interactions create a
novel, informative sensory signal (CNS). In this case it is a straight-line,
vertical motion of one Lego block. It provides evidence in favor of a
prismatic joint in between these two objects (although in this case,
this is still incorrect).}

\BText{When the robot autonomously interacts with the scene it creates these
informative sensory signals not only in the visual but also haptic
sensory modality. This data is strongly correlated with a particular
articulation mechanism. Fig.~\ref{fig:lego} [Right] visualizes this
scenario. By leveraging knowledge of the regularity in $S \times A
\times t$, the robot can also form a correct hypothesis of the
articulation model (APR). The Lego blocks are rigidly attached at first, but
when the robot applies enough vertical force to the top-most Lego block,
there is sensory evidence for a free body articulation model.}

In the literature, there are offline estimation approaches towards
this problem that either rely on fiducial
markers~\cite{sturm2011probabilistic} or 
marker-less tracking~\cite{Pillai-RSS-14,KatzISER2010}. There are also online
approaches~\cite{martin_iros_ip_2014} where the model is
estimated during the movement. Most recent methods include reasoning
about actions to actively reduce the uncertainty in the articulation
model estimates~\cite{hausman15icra, BarraganICRA14,otte2014entropy}.

\subsection{Object Dynamics Learning and Haptic Property Estimation}
\BText{Interactive Perception has also made major inroads into the challenge
of estimating haptic and inertial properties of
objects. Fig.~\ref{fig:ball} shows a simple example scenario that
shall serve to illustrate why
IP simplifies the problem. Consider a sphere that is lying on a
table. The robot is supposed to estimate the 
weight of the sphere given different sources of information. We
assume that the robot knows the relationship between push force,
distance the sphere traveled and sphere weight. In the trivial
static scene scenario illustrated in Fig.~\ref{fig:ball} [Left], the
robot is not able to estimate any of the inertial
properties. It encounters similar problems as in the
previous example (Fig.~\ref{fig:lego}) even if it was able to change
the viewpoint.} 

\BText{In Fig.~\ref{fig:ball} [Center], the robot can observe the motion of the
  sphere that is pushed by a person. Now, the robot can easily segment
  the ball from the table due 
to the additional sensory signal that was not present before
(CNS). However, it remains very difficult for the robot to estimate
the inertial properties of the sphere because it does not know the
strength of the push.  
Without this information, the known regularity in $S \times A \times
t$ cannot be exploited. The robot will only be able to obtain a very
uncertain estimate of the sphere weight because it needs to marginalize
over all the possible forces the person may have applied.}

\BText{In Fig.~\ref{fig:ball} [Right], the robot
  interacts with the sphere. It can control the push force that
  is applied and observe the resulting distance at
  which the sphere comes to rest.  
  Given the knowledge of the strength of the push, it can now exploit the
  known associations between actions and sensory responses to estimate
  the sphere’s inertial properties (APR).}

There are several examples that leverage the insight that IP enables
the estimation of haptic and inertial 
properties. For example, \cite{RomanoTRO12,ChuICRA13} show that surface and
material properties of objects can be more accurately estimated if the
robot's haptic sensor is moved along the surface of the object.

\citet{Atkeson01091986} and \citet{ZhangICRA12}
move the object to estimate its inertial
properties or other parameters of object dynamics which are otherwise
unobservable. 

\subsection{Object Recognition or Categorization}
Approaches to detect object instances or objects of a specific
category have to learn the appearance or shape
of these objects under various conditions. There
are many challenges in object recognition or categorization that make
this task very difficult given only a single input image. A method
has to cope with occlusions, different lighting conditions, scale of
the images, just to name a few. State-of-the-art approaches in
Computer Vision as e.g. \cite{VeryDeep:iclr:2015,NIPS2015_5638} require enormous amounts of
training data to handle these variations.

Interactive Perception approaches allow a robot to move objects 
and hence reveal previously hidden features. Thereby it can resolve
some of the aforementioned challenges autonomously and may alleviate
the need for enormous amounts of training data.
Example approaches that perform object segmentation and
categorization can be found in \cite{SchiebenerICRA14,
  Schiebener2011}. \KText{The challenge of object
recognition/categorization has been tackled by \citet{SinapovICRA14} and
\citet{hausman14iros-ws-b}}.

\subsection{Multimodal Object Model Learning}
Learning models of rigid, articulated and deformable objects is
a central problem in the area of Computer Vision. In the majority of the cases, the model is learned or built from multiple
images of the same object or category of objects. 
Once the model is learned, it can be used to find the object in new,
previously unseen contexts.  

A robot can generate the necessary data through interaction
with the environment. For example,
\citet{Kra11Aut} 
present an approach where a robot autonomously builds an object model while
holding the object in its hand. The object model is completed by
executing actions informed by next best view planning.  
\citet{KenneyICRA09} push an object on the plane and
accumulate visual data to build a model of the object.

There are also approaches that build an object
model from haptic sensory data,
e.g. by~\citet{DragievICRA2011}. \citet{Allen:85,bohg:iros10,IlonenBK14,BjorkmanBHK13} 
show examples that initialize a model from visual data and then further
augment it with tactile data. \citet{SinapovICRA14} present a method where 
a robot grasps, lifts and
shakes objects to build a multi-modal object model.

\subsection{Object Pose Estimation}\label{sec:pose}
Interactive perception has also been applied to the problem of object
pose estimation. Related approaches focus on reducing object pose
uncertainty by either touching or moving it.  

\citet{KovalIROS13} employ manifold particle filters for this
purpose. \citet{JavdaniICRA2013} use information-theoretic criteria
such as information gain to actively reduce the uncertainty of the
object pose. In addition to reducing uncertainty, they also provide
optimality guarantees for their policy. 

\subsection{Grasp Planning}
Cluttered scenes and premature object interactions used to be considered as obstacles for
grasp planning that had to be avoided by all means. In contrast, Interactive Perception approaches in this
domain take advantage of the robot's ability to
move objects out of the way or to explore them to create more
successful plans even in clutter or under partial information.

\KText{\citet{HsiaoICRA2009} use proximity sensors to
estimate the local surface orientation to select a good grasp.}
\citet{DragievICRA2013} devise a grasp controller for objects of
unknown shape which combines both exploration and exploitation
actions. Object shape is represented by a Gaussian Process implicit
surface. Exploration of the shape is performed using tactile
sensors on the robot hand. Once the object model is
sufficiently well known, the hand does not
prematurely collide with the real object during grasping attempts.

\subsection{Manipulation Skill Learning}
In some cases the goal of Interactive Perception is to accomplish a particular manipulation skill. 
This manipulation skill is generally a combination of some of the pre-specified goals. 

To learn a manipulation skill~\citet{Kappler-RSS-15,pastor2011online} represent the task as a sequence of demonstrated behaviors encoded in a manipulation graph. 
This graph provides a strong prior on how the actions should be sequenced to accomplish the task.
\citet{LeeICRA15} uses a set of kinesthetic demonstrations to learn the right variable-impedance control strategy.
\citet{Cusumano-TownerSMOA11} propose a planning approach that uses a previously-learned Hidden Markov Model to fold clothes. 

The approaches discussed above can be thought of as methods that
capture the regularity of complex manipulation behaviors in $S\times A \times t$ by learning them via demonstration.

\subsection{State Representation Learning}
In the majority of the IP approaches, the representation of sensory
data and latent variables are pre-specified based on prior knowledge
about the system and task.
There are however some approaches that learn these
representations. Most notable are
\citet{Jonschkowski-14-AR,LevineFDA15,AssaelWSD15}.
\BText{All of them learn some mapping from raw, high-dimensional sensory
input (in this case images) to a lower-dimensional state
representation. All of these example approaches fix the
structure of this mapping, e.g. linear
mapping with task-specific regularizers~\citep{Jonschkowski-14-AR} or {\em Convolutional Neural
  Networks\/}~\citep{LevineFDA15,AssaelWSD15}. 
  The parameters of this mapping are learned from data.}

 \section{Taxonomy of Interactive Perception}
\label{sec:definitions}
In this section, we identify a number of important aspects that
characterize existing IP approaches. \JText{These are additional to the
  two benefits of CNS and APR and independent of the specific
  application of an approach.} \BText{ We use these aspects to taxonomize and group
approaches in the Tables~\ref{tab:first_table}
and~\ref{tab:second_table}.}
\BText{ In the following, each table column is described in detail in a subsection along with example
approaches. We use {\em paper sets\/} to refer to groups of similar
approaches that address the same application, e.g. either object
segmentation or manipulation skill learning. We split paper sets
further into approaches that either exploit CNS or APR. 
We also list papers separately that do not pursue a unique goal,
e.g. they perform both Object Segmentation and Recognition.}

\begin{figure}[h]
  \begin{center}
    \includegraphics[width=1.0\columnwidth]{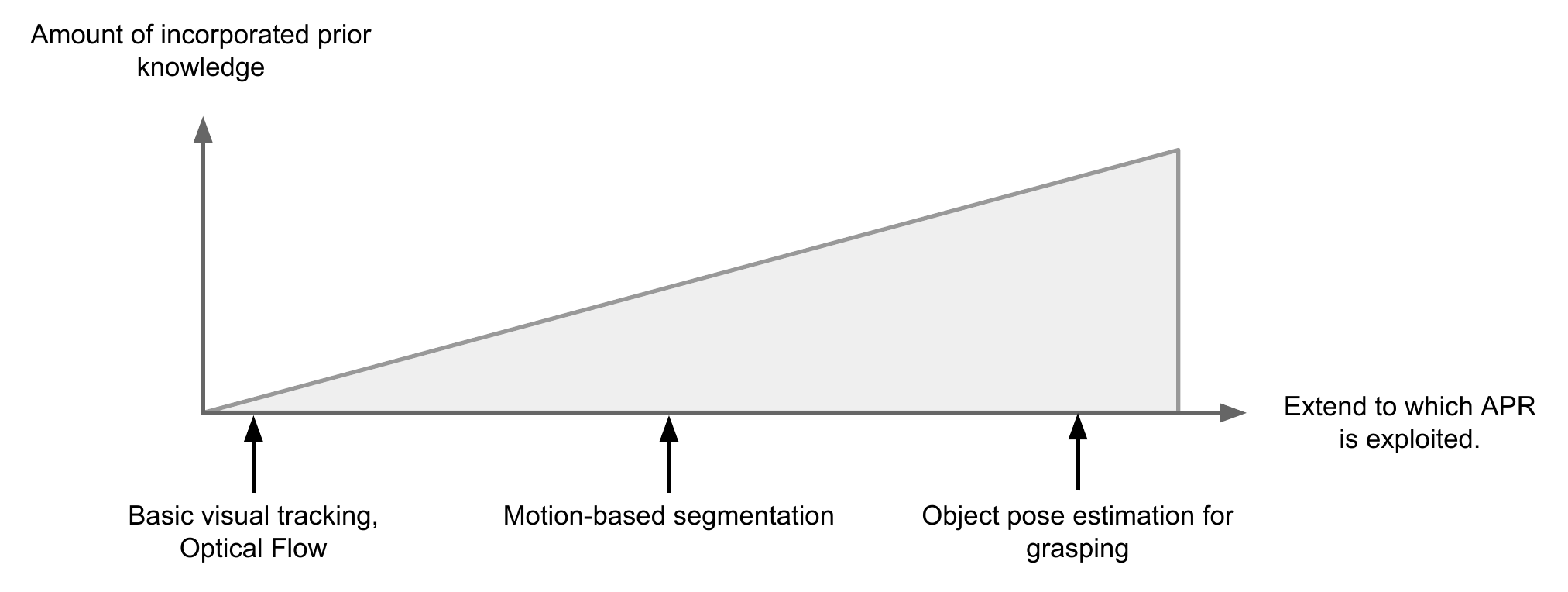}
  \end{center}
  \caption[caption]{\BText{Spectrum of the extent to which knowledge about the Action
    Perception Regularity (APR) is exploited by IP approaches.
    Example problems are plotted along the $x$-axis. Their placement
    depends on how much prior knowledge about the interaction in the
    environment is commonly used in existing approaches towards them.\\
    For example, approaches towards basic visual object tracking or
    optical flow use very weak priors to regularize the solution space
    without incorporating knowledge about the specific interaction
    that caused the novel sensory response (CNS). Similar to that, approaches
    towards motion-based object segmentation often rely on
    interpreting a novel sensory response caused by an arbitrary
    interaction (CNS). Approaches towards object pose estimation often
    choose an action that is expected to provide the most informative
    sensory signal (APR).\\
    The wedge-shape of the graph indicates that the more an approach
    exploits APR, the more prior knowledge it relies
    on. It also indicates that a strict classification of an  
    approach may not be possible in every case.}}
  \label{fig:wedge}
\end{figure}

\begin{sidewaysfigure*}
  \renewcommand{\thefootnote}{\arabic{footnote}}
    \centering\arraybackslash  \rowcolors{2}{gray!25}{white}
   \begin{tabular}[pos]{ | p{1.8cm}| p{1.5cm} | p{2.0cm} | p{2.5cm} | p{2.5cm} | p{2.5cm} | p{3.8cm} | p{5.0cm} |}    
      \hline
      \rowcolor{gray!40}

    {\small\BText{Goals \& Paper Set ID}}&\small Papers &{\small How is the signal in $S
      \times A \times t$ leveraged?~\footnote{Create Novel Signals (CNS) vs Action Perception Regularity (APR)}} &{\small What
      priors are employed?~\footnote{A prior is a source of information that aids in the interpretation of the sensor signal by rejecting noise, possibly by projecting the signal into a lower dimensional space. RO-rigid objects, AP-action primitives, PM-planar motion, OD-object database, SD-simple dynamics, AO-articulated objects, DO-deformable objects.} } 
    &{\small Does the approach perform action selection?~\footnote{Alternatively, it can rely on some hard-coded action or just interpret/exploit the interaction induced by something/someone else. M=myopic/greedy, PH=variable planning horizon, G=global policy}} &
    {\small What is the objective: perception, manipulation or both?} & {\small
      Are multiple sensor modalities exploited?~\footnote{Does the approach use multiple modalities? If so, which ones?}} & {\small
      How is uncertainty modeled and used?~\footnote{Is uncertainty explicitly represented? How is it
  used?  DDM-deterministic dynamics model, SDM-stochastic dynamics
  model, DOM-deterministic observation model, SOM-stochastic
  observation model, EU-estimates uncertainty}} \\    
      
          \tiny{\BText{Object Recognition}}  &\small{\cite{hausman14iros-ws-b, Kjellstrom:CVIU:2010}},\JText{\cite{Cai-RSS-16}} &\small APR &\small RO~\cite{hausman14iros-ws-b}, OD, AP &\small M~\cite{hausman14iros-ws-b},
   \xmark \cite{Kjellstrom:CVIU:2010},\JText{\cite{Cai-RSS-16}} &\small P\cite{Kjellstrom:CVIU:2010, hausman14iros-ws-b},\JText{\cite{Cai-RSS-16}}&\small \xmark (Vision)&\small
   no dynamics~\JText{\cite{Cai-RSS-16}}, SDM~\cite{hausman14iros-ws-b, Kjellstrom:CVIU:2010}, SOM~\cite{Kjellstrom:CVIU:2010, hausman14iros-ws-b},\JText{\cite{Cai-RSS-16}}, EU~\cite{hausman14iros-ws-b, Kjellstrom:CVIU:2010} \\   
 
    \tiny{\BText{Object Segmentation I}}& \small{\cite{FitzpatrickIROS12, KenneyICRA09,MettaNN03, BergstromICVS11, xu_siga15, ChangICRA12, GuptaICRA12, KarolICRA13, BerschRSS-W12, KuzmicHUMANOIDS10, SchiebenerICRA14}, \BText{\cite{PajarinenICRA15}}} & \small CNS &\small
    RO, PM~\cite{MettaNN03,FitzpatrickIROS12, KenneyICRA09,ChangICRA12, GuptaICRA12, KarolICRA13, BerschRSS-W12, KuzmicHUMANOIDS10, SchiebenerICRA14},\BText{\cite{PajarinenICRA15}}, OD~\cite{MettaNN03}, AP~\cite{BergstromICVS11, xu_siga15,ChangICRA12, GuptaICRA12, KarolICRA13, BerschRSS-W12, KuzmicHUMANOIDS10, SchiebenerICRA14} &\small 
    \xmark~\cite{FitzpatrickIROS12, KenneyICRA09},M~\cite{MettaNN03, BergstromICVS11, xu_siga15,ChangICRA12, GuptaICRA12, KarolICRA13, BerschRSS-W12, KuzmicHUMANOIDS10, SchiebenerICRA14}, \BText{\cite{PajarinenICRA15}} 
    &\small P~\cite{FitzpatrickIROS12, KenneyICRA09,MettaNN03, BergstromICVS11, xu_siga15,ChangICRA12, GuptaICRA12, KarolICRA13, BerschRSS-W12, KuzmicHUMANOIDS10, SchiebenerICRA14},M~\BText{\cite{PajarinenICRA15}} &\small \xmark (Vision) & \small
    no dynamic model, DOM~\cite{FitzpatrickIROS12,MettaNN03,ChangICRA12, GuptaICRA12, KarolICRA13, BerschRSS-W12, KuzmicHUMANOIDS10, SchiebenerICRA14}, SOM~\cite{KenneyICRA09,BergstromICVS11, xu_siga15},\BText{\cite{PajarinenICRA15}}, \KText{EU~\cite{xu_siga15},\BText{\cite{PajarinenICRA15}}}\\ 
    
    
    \tiny{\BText{Object Segmentation II}} &\small{\cite{HoofIROS12, hoofHUMANOIDS13}} &\small APR &\small AP &\small M & \small P &\small \xmark (Vision) &\small SDM, SOM, EU \\          
    \tiny{\BText{Object Segmentation - Object Recognition I}} &\small{\cite{Schiebener2011}} &\small CNS &\small RO, AP, PM &\small M &\small P &\small \xmark (Vision) &\small no dynamic model, DOM \\ 
    \tiny{\BText{Object Segmentation - Object Recognition II}}&\small {\cite{UdeIJHR08}} &\small APR & \small RO, AP &\small G &\small P &\small \xmark (Vision) &\small DDM, SOM, EU \\         
    \tiny{\BText{Articulation Model Estimation - Object Segmentation}} &\small\KText{\cite{KatzICRA2011}} &\small \KText{CNS} &\small \KText{AO} & \small\KText{\xmark} &\small \KText{P} &\small \KText{\xmark (Vision)} &\small \KText{no dynamics, SOM}\\  
    \tiny{\BText{Articulation Model Estimation I}}&\small {\cite{BarraganICRA14},\KText{\cite{KatzISER2010, Pillai-RSS-14, martin_iros_ip_2014, sturm2011probabilistic, martinmartin_hoefer_iros_2016}}} &\small CNS &\small AO &\small \xmark~\KText{\cite{KatzISER2010, Pillai-RSS-14, martin_iros_ip_2014, sturm2011probabilistic, martinmartin_hoefer_iros_2016}}, M~\cite{BarraganICRA14} &\small P &\small \xmark (Vision) &\small
    no dynamics~\KText{\cite{KatzISER2010}}, SOM, SDM~\KText{\cite{Pillai-RSS-14, martin_iros_ip_2014, sturm2011probabilistic, martinmartin_hoefer_iros_2016}},\cite{BarraganICRA14}, EU~\KText{\cite{KatzICRA2011, KatzISER2010, Pillai-RSS-14, martin_iros_ip_2014, sturm2011probabilistic, martinmartin_hoefer_iros_2016}}, \cite{BarraganICRA14}\\  
    \tiny{\BText{Articulation Model Estimation II}}& \small{\cite{jain2010pulling, Karayiannidis2013, brock2009learning, hausman15icra, otte2014entropy}} &\small APR &\small AO, PM~\cite{brock2009learning} &\small M~\cite{hausman15icra,otte2014entropy}, G~\cite{jain2010pulling, Karayiannidis2013,brock2009learning} &\small P~\cite{hausman15icra,otte2014entropy}, B~\cite{jain2010pulling, Karayiannidis2013,brock2009learning} &\small
    \cmark  (Vision, Tactile)~\cite{hausman15icra},  (Force/Torque or Joint Positions, Visual Odometry~\cite{jain2010pulling})~\cite{jain2010pulling, Karayiannidis2013}, \xmark(Vision)~\cite{brock2009learning, otte2014entropy} &\small SDM~\cite{hausman15icra, otte2014entropy}, SOM~\cite{hausman15icra, otte2014entropy}, EU~\cite{hausman15icra, otte2014entropy}, DDM~\cite{jain2010pulling, Karayiannidis2013, brock2009learning}, DOM~\cite{jain2010pulling, Karayiannidis2013, brock2009learning} \\ 
     \tiny{\BText{Pose Estimation}}&\small{\cite{JavdaniICRA2013, Mason1990}},\KText{\cite{KovalIROS13}} &\small APR &\small RO, OD, PM~\cite{Mason1990},\KText{\cite{KovalIROS13}}, SD ~\KText{\cite{KovalIROS13}} &\small \xmark~\KText{\cite{KovalIROS13}}, M~\cite{JavdaniICRA2013}, G~\cite{Mason1990} &\small P~\cite{JavdaniICRA2013},\KText{\cite{KovalIROS13}}, M~\cite{Mason1990} &\small \xmark (Tactile~\cite{JavdaniICRA2013},\KText{\cite{KovalIROS13}}), (Vision~\cite{Mason1990}) &\small static environment~\cite{JavdaniICRA2013}, SDM~\cite{Mason1990},\KText{\cite{KovalIROS13}}, SOM~\cite{JavdaniICRA2013},\KText{\cite{KovalIROS13}}, DOM~\cite{Mason1990}, EU~\cite{JavdaniICRA2013},\KText{\cite{KovalIROS13}}\\
     
        \tiny{\BText{Pose Estimation - Object Dynamics Learning I}}
   &\small{\cite{NIPS2015_Galileo}} &\small CNS&\small RO, PM,
   OD&\small \xmark &\small P &\small \xmark
   (Vision)&\small SDM, EU,
   DOM\\ 
   
   \tiny{\BText{Pose Estimation - Object Dynamics Learning II}}
   &\small{\cite{ZhangICRA12}} &\small APR &\small RO, PM, AP &\small \xmark &\small P &\small \cmark (Vision,
   Tactile&\small SDM, EU,SOM\\       
   
   \tiny{\BText{Object Dynamics Learning}}&{ \cite{Atkeson01091986}} &\small APR &\small RO, AP, OD &\small \xmark &\small P &\small \xmark (Force/Torque) &\small DDM, DOM\\

   \tiny{\BText{Grasp Planning I}}&\small{\cite{DogarICRA13}, \JText{\cite{natale2006sensitive}}, \BText{\cite{LevinePKQ16}}} &\small CNS & \small
   RO~\cite{DogarICRA13},\JText{\cite{natale2006sensitive}}, OD~\cite{DogarICRA13},\JText{\cite{natale2006sensitive}}, AP~\cite{DogarICRA13},\JText{\cite{natale2006sensitive}}, \BText{DO~\cite{LevinePKQ16}} &\small
   \xmark \JText{\cite{natale2006sensitive}}, G \BText{\cite{LevinePKQ16}},(dependent on the algorithm in~\cite{DogarICRA13}) & \small
   P~\cite{DogarICRA13}, M~\JText{\cite{natale2006sensitive}},\BText{\cite{LevinePKQ16}} & \small\xmark (Vision) &\small 
   no dynamics, \JText{DOM\cite{natale2006sensitive}}, SOM~\cite{DogarICRA13},\BText{\cite{LevinePKQ16}}, EU~\BText{\cite{LevinePKQ16}}\\   
     \hline                
   \end{tabular}
   \caption{Taxonomy of Interactive Perception approaches - Part 1}
   \label{tab:first_table}
   \end{sidewaysfigure*}

   \begin{sidewaysfigure*}
    \centering\arraybackslash \rowcolors{0}{gray!25}{white}
  \begin{tabular}[pos]{ | p{1.8cm}| p{1.5cm} | p{2.0cm} | p{2.5cm} | p{2.5cm} | p{2.5cm} | p{3.8cm} | p{5.0cm} |}    
    \hline
	\rowcolor{gray!40}
            {\small\BText{Goals \& Paper Set ID}}&\small Papers &{\small How is the signal in $S
      \times A \times t$ leveraged?~\footnote{Create Novel Signals (CNS) vs Action Perception Regularity (APR)}} &{\small What
      priors are employed?~\footnote{A prior is a source of information that aids in the interpretation of the sensor signal by rejecting noise, possibly by projecting the signal into a lower dimensional space. RO-rigid objects, AP-action primitives, PM-planar motion, OD-object database, SD-simple dynamics, AO-articulated objects, DO-deformable objects.} } 
    &{\small Does the approach perform action selection?~\footnote{Alternatively, it can rely on some hard-coded action or just interpret/exploit the interaction induced by something/someone else. M=myopic/greedy, PH=variable planning horizon, G=global policy}} &
    {\small What is the objective: perception, manipulation or both?} & {\small
      Are multiple sensor modalities exploited?~\footnote{Does the approach use multiple modalities? If so, which ones?}} & {\small
      How is uncertainty modeled and used?~\footnote{Is uncertainty explicitly represented? How is it
  used?  DDM-deterministic dynamics model, SDM-stochastic dynamics
  model, DOM-deterministic observation model, SOM-stochastic
  observation model, EU-estimates uncertainty}} \\     
    \tiny{\BText{Grasp Planning II}}&\small\cite{PlattISRR11, kroemer2010combining},\KText{\cite{KovalRSS14, hsiao2011robust,HsiaoICRA2009,Boularias_AAAI_2015, pinto2015supersizing}},\BText{\cite{MARICRA15,PajarinenIROS14}},\JText{\cite{Salganicoff:91}} &\small APR & \small
   RO, AP~\cite{PlattISRR11,Boularias_AAAI_2015},\JText{\cite{Salganicoff:91}},\KText{\cite{pinto2015supersizing}}, \BText{\cite{MARICRA15,PajarinenIROS14}, PM~\cite{MARICRA15}},\KText{\cite{KovalRSS14, hsiao2011robust}},OD~\cite{kroemer2010combining},\JText{\cite{Salganicoff:91}},\KText{\cite{KovalRSS14, hsiao2011robust}}, SD~\KText{\cite{KovalRSS14, hsiao2011robust}} & \small
   \BText{\xmark\cite{MARICRA15}},G~\cite{HsiaoICRA2009,kroemer2010combining},\KText{\cite{KovalRSS14}},  M~\cite{Boularias_AAAI_2015}\KText{\cite{pinto2015supersizing}}\JText{\cite{Salganicoff:91}}, PH~\cite{PlattISRR11},\BText{\cite{PajarinenIROS14}},\KText{\cite{hsiao2011robust}}  & \small
   B~\cite{HsiaoICRA2009},M~\cite{Boularias_AAAI_2015,PlattISRR11, kroemer2010combining}\KText{\cite{pinto2015supersizing}}\JText{\cite{Salganicoff:91}}, \BText{\cite{MARICRA15}} & \small
   \xmark (Proximity Sensors~\cite{HsiaoICRA2009}), (Vision~\cite{Boularias_AAAI_2015},\BText{\cite{MARICRA15,PajarinenIROS14}},\KText{\cite{pinto2015supersizing}}), (Tactile~\KText{\cite{KovalRSS14, hsiao2011robust}}), \cmark (Vision, Tactile)~\cite{PlattISRR11},\JText{\cite{Salganicoff:91}}, (Vision, Joint Ang.)~\cite{kroemer2010combining} &\small
   no dynamic model\cite{HsiaoICRA2009,kroemer2010combining},\BText{\cite{MARICRA15}},\KText{\cite{pinto2015supersizing}}, \JText{\cite{Salganicoff:91}}, DOM\cite{Boularias_AAAI_2015},\BText{\cite{MARICRA15}},\KText{\cite{pinto2015supersizing}}, SOM\cite{HsiaoICRA2009,PlattISRR11, kroemer2010combining}\KText{\cite{KovalRSS14, hsiao2011robust}},\BText{\cite{PajarinenIROS14}},\JText{\cite{Salganicoff:91}}, 
   SDM\cite{Boularias_AAAI_2015,PlattISRR11},\BText{\cite{PajarinenIROS14}},\KText{\cite{KovalRSS14, hsiao2011robust}}, EU~\cite{PlattISRR11, kroemer2010combining},\BText{\cite{PajarinenIROS14}},\KText{\cite{HsiaoICRA2009,Boularias_AAAI_2015, pinto2015supersizing}},\KText{\cite{KovalRSS14, hsiao2011robust}},\JText{\cite{Salganicoff:91}}\\

   
   \tiny{\BText{Grasp Planning - Pose Estimation}}&\small {\cite{LPKTLPICRA12}} &\small APR &\small RO, AP &\small PH &\small M & \small\cmark (Vision, Tactile) &\small SDM, SOM, EU\\     
    
    \tiny{\BText{Haptic Property Estimation I}}&\small { \cite{RomanoTRO12, CulbertsonUK14,ChuICRA13}} & \small CNS &\small RO~\cite{RomanoTRO12, CulbertsonUK14}, OD, PM, AP~\cite{ChuICRA13}  &\small \xmark &\small P &\small \xmark (Tactile) &\small no dynamics, DOM~\cite{RomanoTRO12, CulbertsonUK14}, SOM~\cite{ChuICRA13}\\ 
    \tiny{\BText{Haptic Property Estimation II}}&\small{\JText{\cite{LoebFishel:2014}}} & \JText{APR} &\small \JText{PM, AP, OD} &\small \JText{M} &\small \JText{P} &\small \JText{\xmark (Tactile)} &\small  \JText{no dynamics model, SOM, EU}\\     
        \tiny{\BText{Multimodal Object Model Learning I}}&\small  \cite{KroemerTRO11, IlonenBK14},\JText{\cite{Metta:HapticRepr:04}} &\small CNS &\small RO, AP, PM~\cite{KroemerTRO11}  &\small \xmark &\small P &\small
    \cmark (Vision, Tactile)~\cite{KroemerTRO11, IlonenBK14},\JText{\cite{Metta:HapticRepr:04}} &\small
    no dynamic model~\cite{KroemerTRO11}, \JText{\cite{Metta:HapticRepr:04}}, SDM~\cite{IlonenBK14},\JText{\cite{Metta:HapticRepr:04}},SOM~\cite{KroemerTRO11, IlonenBK14},\JText{\cite{Metta:HapticRepr:04}}, EU~\cite{IlonenBK14}\\ 
    
    \tiny{\BText{Multimodal Object Model Learning II}}   &\small {\cite{DragievICRA2011,Kra11Aut,bohg:iros10,BjorkmanBHK13}},\BText{\cite{NataleBrain07, BrowatzkiTRO14,TsudaARSO11,TorresJara2005}}, \JText{\cite{KraftIJHR08}} &\small APR & \small
    RO,AP~\BText{\cite{NataleBrain07,TorresJara2005}},\cite{bohg:iros10,BjorkmanBHK13},\JText{\cite{KraftIJHR08}}, OD~\BText{\cite{BrowatzkiTRO14,TsudaARSO11}}, PM~\cite{bohg:iros10,BjorkmanBHK13} & \small
    G\cite{DragievICRA2011}, M\cite{Kra11Aut},\BText{\cite{BrowatzkiTRO14,TsudaARSO11}},\xmark~\cite{bohg:iros10,BjorkmanBHK13},\BText{\cite{NataleBrain07,TorresJara2005}},\JText{\cite{KraftIJHR08}} & \small
    P\cite{Kra11Aut,bohg:iros10},\JText{\cite{KraftIJHR08}},\BText{~\cite{NataleBrain07,BrowatzkiTRO14,TorresJara2005}}, B\cite{DragievICRA2011},\BText{\cite{TsudaARSO11}} & \small
    \cmark (Tactile, Vision\cite{DragievICRA2011,bohg:iros10,BjorkmanBHK13},\BText{\cite{NataleBrain07}}), \xmark (Vision~\cite{Kra11Aut},\BText{~\cite{TsudaARSO11}},\JText{~\cite{KraftIJHR08}}),
    \BText{~\cmark(Vision, Joint Ang.~\cite{BrowatzkiTRO14})}, \BText{\cmark (Audio, FT\cite{NataleBrain07})}, \BText{(Tactile, Audio)~\cite{TorresJara2005}} &\small
    static environment~\cite{DragievICRA2011}, no dynamics\cite{Kra11Aut,bohg:iros10,BjorkmanBHK13},\BText{\cite{BrowatzkiTRO14, TsudaARSO11,TorresJara2005}},\JText{\cite{KraftIJHR08}}, \BText{DDM\cite{NataleBrain07}}, \BText{DOM\cite{TorresJara2005}}
    SOM~\cite{DragievICRA2011,Kra11Aut,bohg:iros10,BjorkmanBHK13},\BText{\cite{TsudaARSO11}}, EU~\cite{DragievICRA2011,Kra11Aut,bohg:iros10,BjorkmanBHK13},\BText{\cite{BrowatzkiTRO14}}, \BText{No observation model or EU~\cite{TsudaARSO11}}  \\  
    
    \tiny{\BText{Multimodal Object Model Learning - Object Recognition}}&\small {\cite{SinapovICRA14, SinapovICRA13, SinapovRAS14}, \BText{\cite{Omrcenhumanoids07}}} \JText{\cite{MichelZabulisArgyros2014a}} &\small CNS &\small RO, AP\cite{SinapovICRA14, SinapovICRA13, SinapovRAS14},\BText{\cite{Omrcenhumanoids07}} &\small \xmark &\small P &\small \cmark (Vision, Audio, Tactile), \BText{\xmark (Vision~\cite{Omrcenhumanoids07, MichelZabulisArgyros2014a})} &\small no dynamics\cite{SinapovICRA14, SinapovICRA13, SinapovRAS14}\BText{\cite{Omrcenhumanoids07}}, \JText{SDM\cite{MichelZabulisArgyros2014a}}, SOM\cite{SinapovICRA14, SinapovICRA13, SinapovRAS14}\BText{\cite{Omrcenhumanoids07}},\JText{\cite{MichelZabulisArgyros2014a}}, EU\cite{SinapovICRA14, SinapovICRA13, SinapovRAS14}\BText{\cite{Omrcenhumanoids07}}\\ 
    
    \tiny{\BText{Multimodal Object Model Learning - Grasp Planning}}   &\small {\cite{DragievICRA2013}} &\small APR &\small RO &\small G &\small M &\small \cmark (Tactile, Vision) &\small static environment, SOM, EU \\  
    
    
    \tiny{\BText{Manipulation Skill Learning}}&\small {\cite{Kappler-RSS-15, pastor2011online, LevineICRA15, HanIROS15, Cusumano-TownerSMOA11, LeeICRA15}} &\small APR &\small
    OD, RO~\cite{Kappler-RSS-15, pastor2011online, LevineICRA15, HanIROS15}, DO~\cite{Cusumano-TownerSMOA11, LeeICRA15}, AP~\cite{Cusumano-TownerSMOA11} &\small G~\cite{Kappler-RSS-15, pastor2011online, LevineICRA15, HanIROS15, LeeICRA15}, M~\cite{Cusumano-TownerSMOA11} &\small
    M &\small \xmark (Vision)~\cite{Cusumano-TownerSMOA11}, \cmark (Vision, Internal torques)~\cite{LeeICRA15}, \cmark (Vision, Tactile)~\cite{Kappler-RSS-15, pastor2011online}, (Joint Pos.)~\cite{LevineICRA15, HanIROS15} &\small 
    no dynamics model~\cite{Kappler-RSS-15, pastor2011online}, SDM~\cite{LevineICRA15, HanIROS15,Cusumano-TownerSMOA11, LeeICRA15}, SOM~\cite{Kappler-RSS-15,Cusumano-TownerSMOA11, LeeICRA15}, DOM~\cite{pastor2011online, LevineICRA15, HanIROS15}, EU~\cite{Cusumano-TownerSMOA11, LeeICRA15} \\
    
    \tiny{\BText{Manipulation Skill Learning- State Representation Learning}}&\small {\cite{AssaelWSD15,LevineFDA15}} &\small APR &\small OD, RO &\small G, PH (MPC)~\cite{AssaelWSD15} &\small M &\small \xmark (Vision)~\cite{AssaelWSD15}, (Vision, Joint Pos.)~\cite{LevineFDA15}  &\small DDM~\cite{AssaelWSD15}, SDM~\cite{LevineFDA15}, DOM\\
    
    \tiny{\BText{State Representation Learning}}&\small \KText{\cite{Jonschkowski-14-AR, pinto2016curious}} &\small \KText{APR} &\small \KText{SD~\cite{Jonschkowski-14-AR}, AP, PM, RO} &\small \KText{\xmark} &\small \KText{P} &\small \KText{\xmark (Vision)~\cite{Jonschkowski-14-AR}, \cmark(Vision, Tactile)~\cite{pinto2016curious}}  &\small \KText{DDM, DOM}\\ 
    \tiny{\BText{State Representation Learning - Object Dynamics Learning}}&\small \JText{\cite{AgrawalNAML16, finn2016unsupervised}} &\small \JText{APR} &\small \JText{AP, PM, RO, OD~\cite{AgrawalNAML16}} &\small \JText{M~\cite{AgrawalNAML16}}, \KText{\xmark~\cite{finn2016unsupervised}} &\small \JText{M~\cite{AgrawalNAML16}},\KText{P~\cite{finn2016unsupervised}} &\small \JText{\xmark (Vision)} &\small \JText{DDM, DOM}\\
 
   \hline

  \end{tabular}
  \caption{Taxonomy of Interactive Perception approaches - Part 2}
  \label{tab:second_table}
   \end{sidewaysfigure*}

\subsection{How is the signal in $S \times A \times t$ leveraged?}
\label{sec:first_column}
\KText{An IP approach leverages at least one of the two aforementioned
  benefits:} (i) it exploits a novel sensory signal that is due to 
some time-varying, forceful interaction \KText{(CNS)} or (ii) also leverages
prior knowledge about the regularity in the combined space of sensory
data and action parameters over time $S \times A \times t$ for predicting or interpreting
this signal \KText{(APR)}. 

 \subsubsection{\BText{Commonalities and differences between CNS and APR}}
 \label{sec:relation_perception}
\BText{Approaches that exploit the novel sensory signal (CNS) also
  rely on regularities in the sensory response to an interaction.  
In its most basic form, this regularity is usually linked to some
assumed characteristic of the environment that thereby restricts
the expected response of the world to an {\em arbitrary\/} action.
Even more useful to robust perception and manipulation is to also
include prior knowledge about the response to a {\em specific\/}
interaction (APR).} 

\BText{Existing approaches towards IP cover a broad spectrum of how
  the possibilities afforded by this combined space $S \times A \times
  t$ are leveraged as visualized in Fig.~\ref{fig:wedge}. 
  On the one end of the spectrum, there are approaches such as visual
  tracking or optical flow that use very weak priors to regularize the
  solution space while maintaining generality (e.g. brightness
  constancy or local motion).  
  In the middle, there are approaches that heavily rely on the
  regularity in the sensory response to an {\em arbitrary\/} interaction
  (e.g. rigid body dynamics, motion restricted to a plane or smooth
  motion). At the very end of the spectrum, there are approaches 
  that leverage both assumptions about environmental constraints and
  knowledge about the {\em specific\/} interaction, to robustly
  interpret the resulting sensory signal and enable
  perceptually-guided behavior.
  While using this kind of prior knowledge loses generality, it may
  result in more robust and efficient estimation in a robotics scenario
  due to a simplification of the solution space.
  If an approach leverages APR, then it also automatically leverages
  CNS.} 

\subsubsection{Example approaches}
We start with approaches that exploit the informative sensory signal
that is due to some forceful interaction (CNS). For instance,
\citet{FitzpatrickIROS12, KenneyICRA09} ease the task of
visual segmentation and object model learning by making some general
assumptions about the environment and thereby about the possible
responses to an arbitrary interaction performed by the robot. Example
assumptions are that only rigid objects are present in the scene and that
motion is restricted to a plane. Although the interaction
is carried out by a robot, the available proprioceptive information is
not used in the interpretation of the
signal.
\citet{KatzISER2010,sturm2011probabilistic, Pillai-RSS-14,
  martin_iros_ip_2014} aim at understanding the structure of
articulated objects by observing their motion when they are interacted
with. While objects are not restricted to be rigid or to only move in a
plane, they are restricted to be piece-wise rigid and to move according
to some limited set of articulation mechanisms. 
Approaches by \citet{BergstromICVS11, ChangICRA12, GuptaICRA12,
  KarolICRA13, BerschRSS-W12, KuzmicHUMANOIDS10, Schiebener2011} 
devise different heuristics for selecting actions that generate
informative sensory signals. These are used to ease perceptual tasks
such as object segmentation or object model learning. Similar to the
above, none of the potentially available knowledge about interaction
parameters is used to predict their effect.

The aforementioned approaches use vision as the source for
informative sensory signals. \citet{ChuICRA13,CulbertsonUK14}
demonstrate how either unconstrained interactions in a plane or fixed
interaction primitives lead to novel {\em haptic\/} sensory signals to
ease the learning of material properties.  

Other approaches utilize the regularity in $ S \times A \times t$ to a
much larger extent for easing perception and/or manipulation
(APR). For example, \citet{Atkeson01091986} estimate the dynamics
parameters of a robotic arm and the load at the end-effector. This
requires a sufficient amount of arm motion, measurements of joint
torques, angles, velocities and acceleration as well as knowledge of
the arm kinematics. We can only learn the appropriate model that
predicts arm motion from input motor torques if given this prior
information on the structure of the space $S \times A \times t$ and
data from interaction.
\citet{SinapovICRA14},\cite{SinapovRAS14}, \citet{SinapovICRA13} let a
robot interact with a set of objects that are characterized by different 
attributes such as rigid or deformable, heavy or light, slippery or
not. Features computed on the different sensor modalities serve as the
basis to learn object similarity. The authors show that this task is
eased when the learning process is conditioned on joint torques and
the different interaction behaviors. They also use the knowledge of
the interaction in \cite{SinapovICRA13,SinapovRAS14} to correlate
various sensor modalities in the $S\times A \times t$. 

\citet{ZhangICRA12, KovalIROS13} track object pose using visual and
tactile data while a robot is pushing this object on a plane.
\citet{ZhangICRA12} solve a non-linear complementarity problem within
their dynamics model to predict
object motion given the control input. At the same time, they use observations of the object
during interaction for estimating parameters of this model such as the
friction parameters. 
\citet{KovalIROS13} assume knowledge of a
lower-dimensional manifold that describes the 
different contact states between a specific object and hand during a
push motion. Hypotheses about
future object poses are constrained to lie on this manifold. 
\KText{\citet{hausman15icra,hsiao2011robust} condition on the action
  to drive the estimation process. \citet{hsiao2011robust} estimate
  the belief state by conditioning the observations on the expected
  action outcomes.} \citet{hausman15icra} adopt a similar approach to
estimate the distribution of possible articulation models based on
action outcomes. 

\subsection{What priors are employed?}
To devise an IP system means to interpret and/or deliberately generate
a signal in the $S \times A \times t$. The regularity of this signal
can be programmed into the system as a 
prior incorporating knowledge of the task; it can be learned from
scratch or the system can pick up these regularities using a mixture
of both priors and learning. Therefore an important component of any IP system
is this regularity and how it is encoded and exploited for
performing a perception and/or manipulation task.

\subsubsection{Priors on the Dynamics}
Interactive Perception requires knowledge of how actions change the state of the environment. 
Encoding this kind of regularity can be done in a dynamics 
model i.e. the model for predicting the evolution of the environment
after a certain action has been applied. 
Dependent on the number of objects in the environment,
this prediction may be very costly to compute. Furthermore, due
to uncertainty and noise in robot-object and object-object
interactions, the effects of the interactions are stochastic.

\paragraph{Given/Specified/Engineered Priors}
There are many approaches that rely on priors which simplify the
dynamics model and thereby make it less costly to predict the effect
of an action. Examples of commonly used priors are the occurrence of
only rigid objects (RO), of articulated objects (AO) with a discrete
set of links or of only deformable objects (DO). Another prior
includes the availability of a set of action primitives (AP) such as push, pull, grasp, etc. These action primitives
are assumed to be accurately executed without failure. Many approaches
assume that object motion is restricted to a plane (PM) or other
simplifications of the scene dynamics (SD), e.g. quasi-static motion during multi-contact 
interaction between objects. In this section, each prior will be explained in more detail by using one or several
example approaches that exploit them.

Of the highlighted priors some are more commonly used than others. For
instance apart from papers in paper set \BText{(Object Segmentation II, Object Segmentation - Object Recognition II, Haptic Property Estimation II)} almost all other
approaches make assumptions about the nature of objects in the
environment, i.e they assume that all objects present in the environment belong exclusively to one of three classes: rigid, articulated or deformable.

The majority of approaches in Interactive Perception assume
that the objects are rigid (RO). Only approaches concerned with
estimating an articulation model assume the existence of articulated
objects. 
Similarly, \citet{LevinePKQ16,Cusumano-TownerSMOA11, LeeICRA15} in paper set \BText{Manipulation Skill Learning} are unique in that they are the only ones that deal with the manipulation of 
deformable objects (DO).

Many approaches in the paper set \BText{Object Segmentation I} utilize the planar motion prior (PM). In
instances such as \citet{GuptaICRA12}, this prior is used for scene
segmentation where all the objects in the scene 
are assumed to lie on a table plane. In other approaches e.g.
\cite{ChangICRA12, KarolICRA13, BerschRSS-W12, KuzmicHUMANOIDS10,
  SchiebenerICRA14} in \BText{Object Segmentation I},
\cite{KroemerTRO11} in \BText{Multimodal Object Model Learning I} and
\cite{bohg:iros10, BjorkmanBHK13} in \BText{Multimodal Object Model
  Learning II} the planar motion assumption is used not only for scene segmentation, but also to track the movement of objects in the environment.

Then there are approaches which make additional simplifying assumptions about the
dynamics of the system (SD). For instance \citet{KovalRSS14} assume
that the object being manipulated has quasi-static dynamics and moves
only on a plane (PM).
Such an assumption becomes particularly useful in cases where action
selection is performed via a multi-step planning procedure because it
simplifies the forward prediction of object motion.

\paragraph{Learned Priors}
There are approaches that learn a dynamics model of the environment given an
action. Some of these let the robot learn this
autonomously through trial and error. Early approaches towards this are
by~\citet{Mason1990, MettaNN03} that learn a simple mapping from the
current state and action to a most likely outcome. \citep{Mason1990} 
demonstrate this in a tray-tilting task for bringing the object lying on
this tray into a desired configuration. \citep{MettaNN03}
demonstrate their approach in an object pushing behavior and learn
the response of an object to a certain push direction. Both of them
model the non-determinism of the response of the object to an action. 
More recent approaches are presented
by~\citet{LevineICRA15,HanIROS15,AssaelWSD15} where the authors learn
the mapping from current state to next best action in a policy search
framework.
\citet{Kappler-RSS-15,pastor2011online,LeeICRA15} bootstrap the search
process through trial and error by demonstrating actions.



\subsubsection{Priors on the Observations}
Regularities can also be encoded
in the observation model that relates the state of the system
to the raw sensory signals. Thereby it can predict the observation
given the current state estimate. Only
if this relationship is known, an IP robot can gain information from
observations. This information may be about some quantity of interest
that needs to be either estimated or directly provide the distance to some goal
state.
\paragraph{Given/Specified/Engineered Observation Models}
Traditionally, the relationship between the
state and raw sensory signals is hand-designed based on some expert
knowledge. \BText{One example are models of multi-view or perspective
geometry for camera sensors~\citep{Aloimonos88,ByravanF16}.} Often, approaches also
assume access to an object database (OD) that allows them to predict
how the objects will be observed through a given sensor, e.g. by
\citet{ChuICRA13}. 
\paragraph{Learned State Representations}
More recently, we see more approaches that  
learn a suitable, task-specific state representation directly from
observations. Examples
include~\citet{Jonschkowski-14-AR,LevineFDA15,AssaelWSD15} who each
use raw pixel values as input and learn the lower-dimensional
representation jointly with the policy that maps these learned
states to actions. \citep{Jonschkowski-14-AR} achieve this by
introducing a set of hand-defined priors in a loss function that is
minimal if the state representation best matches these
priors. The mapping from raw pixels to the lower dimensional
representation is linear. \citep{LevineFDA15} map the raw pixel values through a
non-linear Convolutional Neural Network (CNN) to a set of feature locations in
the image and initialize the weights for an object pose estimation
task. Both the type of function approximator (CNN encoding receptive
fields) and the data for initialization can be seen as a type of
prior. \citep{AssaelWSD15} use an autoencoder framework where the
authors not only minimize the reconstruction error from the
low-dimensional space back to the original space but also optimize the consistency in the
latent, low-dimensional space.

In the case, where the mapping between state and observation is
hand-designed, the state usually refers to some physical quantity. In the
case where the state representation is learned, it is not so
easily interpretable.

\subsection{Does the approach perform action selection?}
\label{sec:action_selection}
Knowledge about the structure of $S \times A \times t$ can also be exploited to
select appropriate actions.  A good action will reveal as much information as
possible and at the same time bring the system as close as possible to
the manipulation goal.  If we know something about the structure of $S
\times A \times t$, we can perform action selection so as to make the resulting
sensor information as meaningful as possible.  The agent must balance
between exploration (performing an action to improve perception as
much as possible) and exploitation (performing an action that
maximizes progress towards the manipulation goal).

\subsubsection{Problem Formulation}
For optimal action selection, the IP agent needs to know a
policy that given the current state estimate returns the
optimal action or sequence of actions to take. Here, optimal
means that the selected actions yield a maximum expected {\em
  reward\/} to the IP agent. The specific definition of the reward 
function heavily depends on the particular task of the robot. If it is
a purely perceptual task, actions are often rewarded when they reduce the
uncertainty about the current estimate (exploration), e.g. \citet{HoofIROS12}. If the task is a 
manipulation task, actions may be rewarded that bring you closer to a
goal (exploitation), e.g. \citet{LevineICRA15}. 

Finding this policy is one of the core problems for action selection.
Its formalization depends on whether the state of the dynamical system
is directly observable or whether it needs to be estimated from noisy
observations. It also depends on whether the dynamics model is
deterministic or stochastic.


\subsubsection{Dynamics Model}\label{sec:action_select}
Knowing the dynamics model is even more important for action selection
than for improving perception. It allows to predict the effect of an
action on the quantity of interest and thereby the expected reward.
A common way to find the optimal sequence of actions that maximizes
reward under deterministic dynamics is forward or backward value iteration 
\cite{LavallePlanning06}. 

As mentioned 
earlier, a realistic dynamics model should be stochastic to account
for uncertainty in sensing and execution. In this case, to find the
optimal sequence of actions the agent has to form an expectation over
all the possible future outcomes of an action. The dynamical system
can then be modeled as an MDP. Finding the optimal sequence of
actions can be achieved through approaches such as {\em value\/} or
{\em policy\/} iteration~\citep{Bagnell_2013_7451}. 

In an MDP, we assume that the state of the system is
directly observable. However in a realistic scenario, the robot can only
observe its environment through noisy sensors. This can be modeled
with a POMDP where the agent has to maintain a probability
distribution over the 
possible states, i.e. the belief, based on an observation model.
For most real-world problems, it is intractable to find the optimal
policy of the corresponding POMDP. Therefore, there exist many methods
that find approximate solutions to this problem~\citep{Shani:2013}.

\JText{PSRs are another formalism for action selection. Here, the system
dynamics are represented directly by observable quantities in the form
of a set of tests instead of over some latent state representation as
in POMDPs~\citep{Singh:2004,Boots:2011,stork2015learning-A}.}

\subsubsection{Planning Horizon}
Action selection methods can be categorized based on 
the number of steps they look ahead in time.
There are approaches that have a single step look ahead which are
called myopic or greedy (M). Here the agent's actions are optimized
for rewards in the next time step given the current state
of the system. 
Most approaches to
interactive perception that exploit the knowledge of the outcome of an action in
$S \times A \times t$ are myopic (M). Myopic approaches do not
have to cope with the evolution of complex system dynamics or observation models beyond a single step. Hence this considerably reduces the size of the possible solution space.
Examples of such approaches can be seen in paper sets \BText{Object Segmentation II \cite{hausman15icra,otte2014entropy}, in Articulation Model Estimation II \cite{JavdaniICRA2013} and in the paper set Pose Estimation.}

Then, there are approaches which look multiple steps ahead in time to inform their action selection process. 
\BText{These multi-step look-ahead solutions decide an optimal course of
action also based on the current state of the system. The time horizon for these multi-step look-aheads can either be fixed or variable.
In either case, the time horizons are generally dictated by a budget, examples of which include computational resources, uncertainty about the current state, costs associated with the system, etc.
For instance, a popular multi-step look ahead approach relies on the
assumption that the {\it maximum likelihood estimate} (MLE)
observation will be obtained in the future.} 
This way, one can predict the behavior of the system within the time horizon and use it to select an action.
Overall we label such approaches to action selection as planning horizon approaches (PH). 
Examples of these approaches include~\cite{LPKTLPICRA12, PlattISRR11}.



Another set of methods tries to find global policies that specify the
action that should be applied at any point in time for any state. We
categorize such approaches as methods that have global policies (GP)
Among these, there are approaches that take into account all possible
distributions over the state space ({\it beliefs}) and offer globally
optimal policies.  
These policies account for stochastic belief system dynamics, i.e they
maintain probabilities over the possible current states and probable
outcomes given an action.  
Such methods are often solved by formulating them as POMDPs.
In practice the solution to such problems are intractable and are often solved by approximate offline methods. 
\citet{JavdaniICRA2013, KovalRSS14} demonstrate such an approach to action selection for interactive perception.
Another way of finding global policies uses reinforcement learning which provides a methodology to improve a policy over time. 
An example of a specific policy search method is presented by~\citet{LevineICRA15, HanIROS15, LevineFDA15}.

\KText{Apart from planning based approaches that perform action selection, there are approaches that focus on low-level control.
In these approaches, the control input is computed online for the next cycle based on a global control law.
We also classify these methods as global-policy (GP) approaches as they compute
the next control input based on control law that is global, e.g. the feedback matrix in Linear Gaussian Controllers. 
The actions are generated at a high frequency and operate on low-level
control commands.
Examples of these approaches include~\cite{jain2010pulling, Karayiannidis2013, HsiaoICRA2009, DragievICRA2013, DragievICRA2011}.}

\subsubsection{Granularity of Actions}
Action selection can be performed at various granularities. 
For example, a method may either select the next best control input or an entire high-level action.
The next best controls can be low-level motor torques that are
sent to the robot in the next control cycle.
The corresponding action selection loop is executed at a very high
frequency and is dependent on the immediate feedback from different
sensors~\cite{jain2010pulling, Karayiannidis2013, HsiaoICRA2009, 
  DragievICRA2013, DragievICRA2011}.

High-level action primitives are generally used in approaches that do
not require reasoning about fine motor control such as pushing or
grasping actions that are represented by motion primitives. 
In such cases, reasoning about observations is purely dependent on the outcome of high-level actions. 
  There are numerous approaches that utilize high-level actions for interactive perception. 
  Examples include: \citet{BarraganICRA14} and the following authors in paper set \BText{Object Segmentation I}: \citet{FitzpatrickIROS12, MettaNN03, KenneyICRA09} and \citet{BergstromICVS11,sturm2011probabilistic, Pillai-RSS-14,
martin_iros_ip_2014}.


 \subsection{What is the objective: Perception, Manipulation or Both?}
Approaches to Interactive Perception may pursue a perception or a
manipulation goal and in some cases both (see Fig.~\ref{fig:categories}).
Object segmentation, recognition and pose estimation, multi-modal
object model learning and articulation model estimation are examples
of areas where interactive perception is utilized to service
perception. 

Then there are interactive perception approaches whose primary
objective is to achieve a manipulation goal (e.g. grasping or learning
manipulation skills). 
For instance \citet{Kappler-RSS-15, pastor2011online}, exploit
regularities in $S \times A \times t$  to enable better action
selection. 
The robot compares the observed perceptual signal with the expected
perceptual signal given the current manipulation primitive. It then
picks controls that drive the system towards
the expected signal. 
Similarly, \citet{KovalIROS13, LPKTLPICRA12, PlattISRR11}
exploit the regularities in $S \times A \times t$ to facilitate task
oriented grasping, i.e locate and grasp an object of
interest. 

The final thread of interactive perception approaches include a
combination of both perception and manipulation. 
For instance, \citet{DragievICRA2013,KovalRSS14} simultaneously improve perception (object model 
reconstruction or pose estimation, respectively) and select better
actions under uncertainty (efficient grasping).
In \citet{jain2010pulling, Karayiannidis2013} in
paper set \BText{Articulation Model Estimation II}, the knowledge about the
regularity in both the observations and dynamics in $S \times A \times t$ is used to
improve articulation model estimation as well as to enable better
control. In the case of \citet{Karayiannidis2013}, the control input
is directly incorporated into the state estimation procedure. In
contrast, \citet{jain2010pulling} use the position of the end effector
in the articulation mechanism estimation. 
The manipulation goal in both these approaches is to enable a robot to open doors and drawers.


\subsection{Are multiple sensor modalities exploited?}
Some approaches exploit multiple modalities in
the $S\times A \times t$ space, whereas other approaches restrict
themselves to a single informative modality. 
The various sensing modalities can be broadly categorized into contact
and non-contact sensing. Examples of non-contact sensing include
vision, proximity sensors, sonar, etc. Contact
sensing is primarily realized via tactile sensors and force-torque sensors.  Approaches that
only use tactile sensing include the works of
\citet{ChuICRA13,KovalIROS13,KovalRSS14,JavdaniICRA2013}. There are
also approaches that use both 
contact and non-contact sensing to inform the signal in the $S\times
A \times t$ space.
These include some of the works listed in paper sets
\BText{Articulation Model Estimation II, Pose Estimation - Object
  Dynamics Learning II, Multimodal Object Model Learning I \& II and
  Manipulation Skill Learning} in Tables~\ref{tab:first_table}
and~\ref{tab:second_table}. 

\subsection{How is uncertainty modeled and used?}
In Interactive Perception tasks, there are many sources of uncertainty
about the quantity of interest. One of them is the noisy sensors
through which an agent can only partially observe the current state of
the world. Another is the dynamics of the environment in response
to an interaction. Some approaches towards Interactive Perception
model this uncertainty in either their observations and/or the
dynamics model of the system. Depending on their choice, there are a wide variety of 
options for estimating the quantity 
of interest from a signal in $S \times A \times t$. For updating the
current estimate, some approaches use {\em
  recursive state estimation\/} and maintain a full posterior
distribution over the variable of interest,
e.g.~\citep{ZhangICRA12,martin_iros_ip_2014,KovalIROS13}. Others frame 
their problem in terms of energy minimization in a graphical model and
only maintain the {\em maximum a posteriori\/} (MAP) solution from frame to frame,
e.g.~\citep{BergstromICVS11}. An MLE
of the variable of interest is computed in approaches that do not maintain a distribution over possible states.
Examples are clustering methods that assign fixed labels
\cite{ChangICRA12, GuptaICRA12, KarolICRA13, BerschRSS-W12} to the
variable of interest. 
More recently
{\em non-parametric approaches} have also been utilized. For instance~\citet{Boularias_AAAI_2015} use {\it kernel
  density estimation}.
 
 Methods that model uncertainty of the variable(s) of interest can cope better with noisy observations or dynamics, but they become slower to compute as the size of the solution space grows. 
  This creates a natural trade-off between modeling uncertainty and computational speed.
The above choices also have implications for action selection. If we maintain a full distribution over the quantity of interest, then computing a policy that takes the
stochasticity in the dynamics and observation models into account is generally intractable \cite{LavallePlanning06}.
If an approach assumes a known state, the dynamical system can also be
modeled by an MDP with stochastic dynamics given an action. 
The least computationally demanding model for action selection is the
one that neglects any noise in the observations or dynamics. 
However, it might also be the least robust depending on the true
variance in the real dynamical system that the agent tries to
control. 


Based on the above, we propose four labels for IP approaches
with respect to their way of modeling and incorporating uncertainty in
estimation and manipulation tasks.
Approaches that assume deterministic dynamics are labeled (DDM), stochastic dynamics (SDM), deterministic observations (DOM), stochastic observations (SOM) and approaches that estimate uncertainty are labeled (EU). 

\citet{FitzpatrickIROS12, MettaNN03, KenneyICRA09} propose example
approaches that assume no stochasticity in the system, and model both
the dynamics and observations deterministically. Then there are
approaches that assume deterministic observations but do not model the
dynamics at all. These are listed in paper set \BText{Object Segmentation I} which include the
works of \citet{ChangICRA12, GuptaICRA12, KarolICRA13, BerschRSS-W12,
  KuzmicHUMANOIDS10, SchiebenerICRA14}. 
Then there are
approaches that model only stochastic observations but no dynamics
because they assume that the environment is static upon interaction,
e.g. \citet{HsiaoICRA2009}. Most approaches that
assume both stochastic dynamics and observations have some form of
uncertainty estimation technique implemented to account for the
stochasticity in the system. An approach that assumes stochasticity in its observations but does not estimate uncertainty is \citet{ChuICRA13}. Here the authors train a max-margin classifier to assign labels to stochastic observations.

 \section{Discussion and Open Questions}
\label{sec:openQuestions}
\subsection{Remaining Challenges}
If Interactive Perception is about merging perception and manipulation
into a single activity then the natural question arises of how to
balance these components. When have manipulation actions (that are in
service of perception) elicited sufficient information about the world
such that manipulation actions can succeed that are in service
of a manipulation goal? This question bears significant similarities
with the exploration/exploitation trade-off encountered in
reinforcement learning. One can further ask: how can manipulation
actions be found that combine these two objectives---achieving a goal 
and obtaining information---in such a way that desirable criteria
about the resulting sequence of actions (time, effort, risk, etc.) are
optimized? 

When performing manipulation tasks, humans aptly combine different
sources of information, including prior knowledge about the world and
the task, visual information, haptic feedback, and acoustic
signals. Research in Interactive Perception is currently mostly
concerned with visual information. New algorithms are necessary to
extend IP towards a multi-modal framework, where
modalities are selected and balanced so as to maximally inform
manipulation with the least amount of effort, while achieving a
desired degree of certainty. Furthermore, for every sensory channel,
one might differentiate between passively (e.g. just look), actively
(e.g. change vantage point to look), and interactively (e.g. observe
interaction with the world) acquired information. Each of these is
associated with a different cost but also with a different expected
information gain. In addition to adequately mining information from
multiple modalities, Interactive Perception must be able to decide in
which of these different ways the modality should be leveraged.

Also at the lower levels of perception significant changes might be
required.  It is conceivable that existing representations of sensory
data are not ideal for Interactive Perception. Given the focus on
dynamic scenes with multiple moving objects, occlusions, lighting changes, and new objects
appearing and old ones disappearing --- does it make sense to tailor
visual features and corresponding tracking methods to the requirements
of Interactive Perception?  Are there fundamental processing steps,
similar to edge or corner detection, that are highly relevant in the
context of Interactive Perception but have not seen a significant need
in other applications of computer vision?  The same for haptic or
acoustic feedback: when combined with other modalities in the context
of Interactive Perception, what might be the right features or
representations we should focus on? 

\subsection{A Framework for Interactive Perception?}
All of the aforementioned arguments indicate that Interactive
Perception might require a departure from existing perception
frameworks, as they can be found in applications outside of robotics,
such as surveillance, image retrieval, etc.  In Interactive
Perception, manipulation is an integral component of perception. The
perceptual process must continuously trade off multiple sensor
modalities that might each be passive, active, or interactive. There
is no stand-alone perceptual process and not only a single aspect of
the environment that must be extracted from the sensor stream as the
optimization objectives may change when the robot faces different
tasks over its lifetime.  

\BText{After the review of existing work in the field, we conclude that there
  is yet no framework that can address all the challenges in Interactive
  Perception. There are however candidates that represent the regularity in
  $S\times A\times t$ in a way that caters to a particular
  challenge encountered in IP. For instance,
  \citet{Krüger2011740} present a concept that allows to 
  symbolically represent continuous sensory-motor experience:} {\em
  \BText{Object-Action Complexes}\/} \BText{(OACs). The concept's
  current instantiations through the examples in~\citep{Krüger2011740}
  are focused on learning and detecting {\em
    affordances\/}~\citep{Gibson1979} which describe the relationship 
  between a certain situation (often including an object) and the
  action that it allows. }

\BText{Other popular formalisms lend themselves particularly well to the
problem of optimal action selection (see
Section~\ref{sec:action_selection}). Examples include MDPs, POMDPS,
PSRs or Multi-armed bandits. They rely on different
assumptions (e.g. Markov Assumptions, observable state) and make different algorithmic
choices (e.g. probabilistic modeling). Approaches that rely on
these decision-making frameworks often assume the
availability of transition, observation and reward functions and the
possibility to analytically compute the optimal action.}

\BText{For complex real-world problems this is often not the case and
  information about the world can only be collected through
  interaction. The data collected in this way is then used to update
  the relevant models. The problem of selecting the next best action may be
  based on submodularity~\citep{JavdaniICRA2013}, the variance in
  a Gaussian Process~\citep{DragievICRA2011,bohg:iros10} or the
  Bhattacharyya coefficient between two normal distributions
  \citep{FishelLoeb:2012, LoebFishel:2014}.}

\BText{Reinforcement learning~\citep{Bagnell_2013_7451} is also a
  common choice to learn a 
policy for action selection under these complex conditions. Many
approaches assume the availability 
of some reliable state estimator (e.g. by using motion capture or
marker-based systems) where the state is of relatively low dimension
and hand-designed. Particularly relevant to
Interactive Perception are recent approaches that directly learn a
state representation from data and employ reinforcement learning on
this learned state
representation~\citep{LevineFDA15,Jonschkowski-14-AR,AgrawalNAML16}.   
}  

\BText{All these formalisms have been used to solve particular
  subproblems encountered in the context of Interactive
  Perception. We do not claim that this list is complete. However, the
  wealth of very different approaches suggests that there is currently
  not one framework for IP that can address all the relevant
  challenges. It is an open question what such a framework
  would be and how it could enable coordinated progress by developing
  adequate subcomponents.} 

\subsection{New Application Areas}
\BText{The majority of the work that is included in this survey is concerned
with Interactive Perception for manipulating and grasping objects
in the environment. In the context of the recent {\em Darpa Robotics
  Challenge\/} (DRC) we have also seen a need to bridge the gap
between perception and action in whole-body, multi-contact motion
planning and control. The ability to physically explore
unstructured environments (such as those encountered in disaster
sites) are of utmost importance for the safety and robustness of a
robot. Probing and poking not only with your hands but also your legs
can also help extract more information. Currently, these robots
extensively rely on teleoperation and carefully designed user
interfaces~\citep{drc_chris,drc_ihmc}. We argue that they can achieve a
much higher degree of autonomy if they rely on Interactive Perception.}


 \section{Summary}
\label{sec:conclusion}
\BText{This survey paper provides an overview on the current state of the art
in Interactive Perception research. In addition to presenting the 
benefits of IP, we discuss various criteria for categorizing existing work.
We also include a set of problems such as object segmentation,
manipulation skills and object dynamics learning that are commonly
eased using concepts of interactive perception.}

\BText{We identify and define the two main aspects of Interactive
Perception. (i) Any type of forceful interaction with the environment
creates a new type of informative sensory signal that would 
otherwise not be present and (ii) any prior knowledge about the
nature of the interaction supports the interpretation of the signal in
the Cartesian product space of $S \times A \times t$. We use these
two crucial aspects of IP as criteria to include a paper as related or not.
Furthermore, we compare IP to existing perception approaches and named
a few formalisms that allow to capture an IP problem.}

\BText{We hope that this taxonomy helps to establish benchmarks for comparing
various approaches and to identify open problems.}



\section*{Acknowledgment}
The authors would like to thank the anonymous reviewers for their
insightful comments and all the cited authors who provided feedback
upon our request. They would also like to sincerely thank Aleksandra
Waltos for providing the visuals in Figures~\ref{fig:lego}
and~\ref{fig:ball}. 

This research is supported in part by National Science Foundation
grants IIS-1205249, IIS- 1017134, EECS-0926052, the Office of Naval
Research, the Okawa Foundation, and the Max-Planck-Society.  
It is also supported by grant BR 2248/3-1 by the German Science
Foundation (DFG), and grant H2020-ICT-645599 on Soft Manipulation
(Soma) by the European Commission. The authors would also like to
thank Swedish Research Council and Swedish Foundation for Strategic
Research.  
Any opinions, findings, and conclusions or recommendations
expressed in this material are those of the authors and do not
necessarily reflect the views of the funding organizations.

 {\small
\bibliographystyle{./sty/IEEEtranN}
\bibliography{IEEEabrv,bib/TRO15_Bibliography}
}
\begin{IEEEbiography}[{\includegraphics[width=1in,height=1.25in,clip,keepaspectratio]{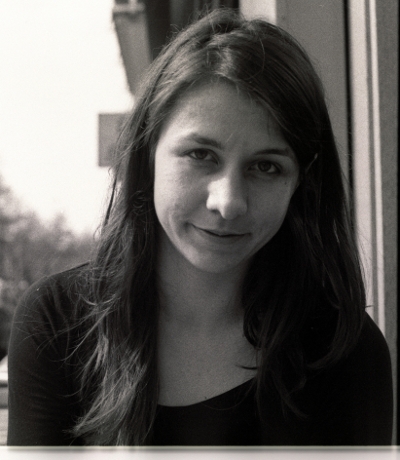}}]
{Jeannette Bohg} is Assistant Professor for Robotics at the Computer Science Department of Stanford University (CA, USA). She is also a guest researcher at the Autonomous Motion Department, Max Planck Institute for Intelligent Systems in T{\"u}bingen, Germany, where she was a research group leader until fall 2017. She holds a Diploma in Computer Science from the Technical University Dresden, Germany and a M.Sc. in Applied Information Technology from Chalmers in G{\"o}teborg, Sweden. In 2012, she received her PhD from the Royal Institute of Technology (KTH) in Stockholm, Sweden.  
Her research interest lies at the intersection between Computer Vision, Robotic Manipulation and Machine Learning. Specifically, she analyses how continuous feedback from multiple sensor modalities help to improve autonomous manipulation capabilities of a robot.  
\end{IEEEbiography}
\begin{IEEEbiography}[{\includegraphics[width=1in,height=1.25in,clip,keepaspectratio]{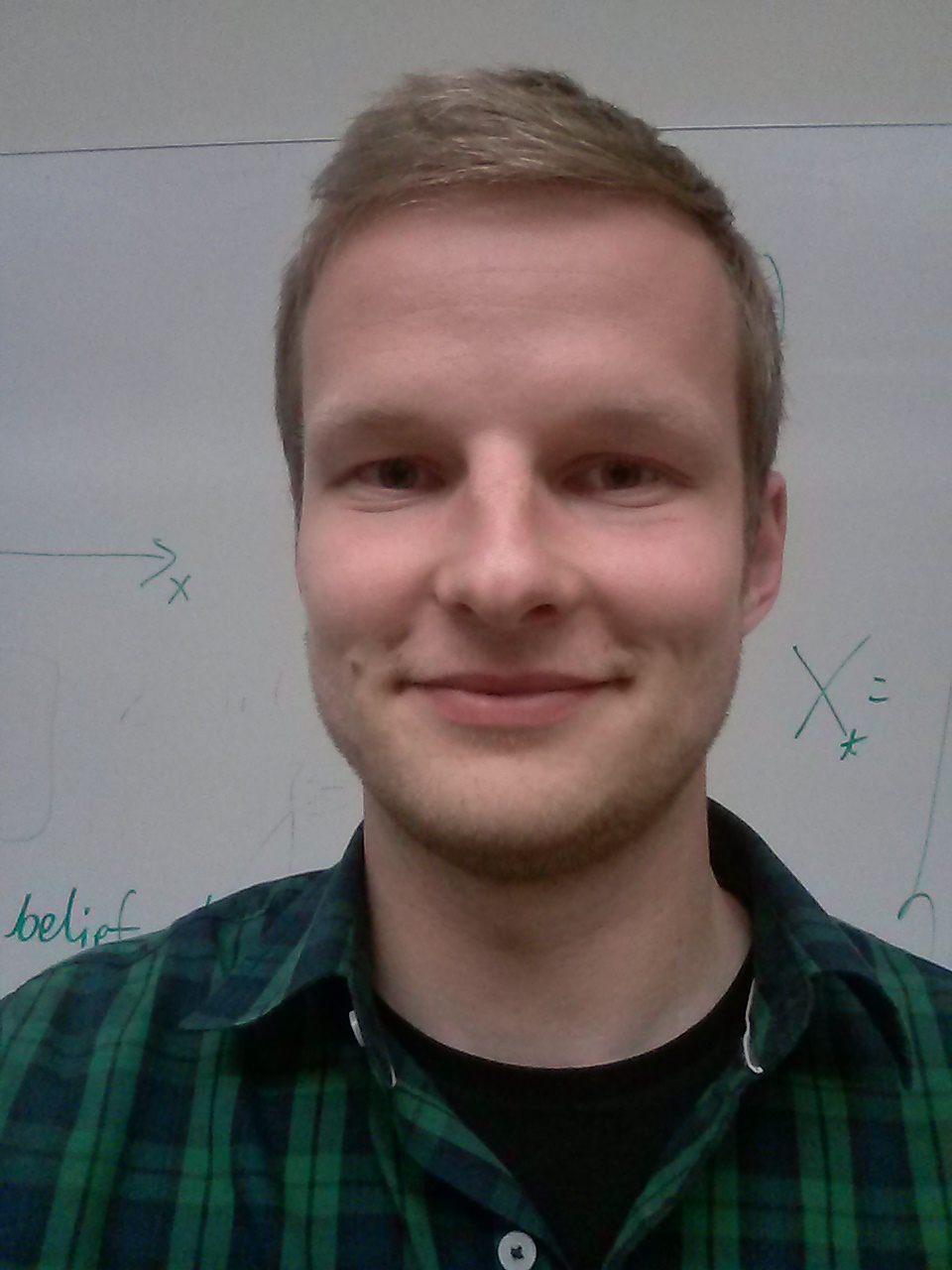}}]
{Karol Hausman}
is a PhD student in computer science at the University of Southern California. He received his M.E. degree in mechatronics from Warsaw University of Technology, Poland, in 2012. In 2013 he graduated with a M.Sc. degree in Robotics, Cognition and Intelligence from Technical University Munich, Germany.
His research interests lie in active state estimation, control generation and machine learning for robotics. More specifically, he is concentrating on model-based and learning-based approaches that address aspects of closing perception-action loops.
\end{IEEEbiography}
\begin{IEEEbiography}[{\includegraphics[width=1in,height=1.25in,clip,keepaspectratio]{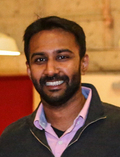}}]%
{Bharath Sankaran}
is a PhD student in computer science at the University of Southern California. He received his B.E degree in mechanical engineering from Anna University, Chennai, India in 2006. He also received an M.E. degree in aerospace engineering from the University of Maryland, College Park, MD, in 2008, an M.S. degree in robotics from the University of Pennsylvania, Philadelphia, PA, in 2012 and an M.S. in Computer Science from the University of Southern California, Los Angeles, CA, in 2015. 
His research interests include applying statistical learning techniques to perception and control problems in robotics. Here he primarily focuses on treating traditional computer vision problems as problems of active and interactive perception.
\end{IEEEbiography}
\begin{IEEEbiography}[{\includegraphics[width=1in,height=1.25in,keepaspectratio]{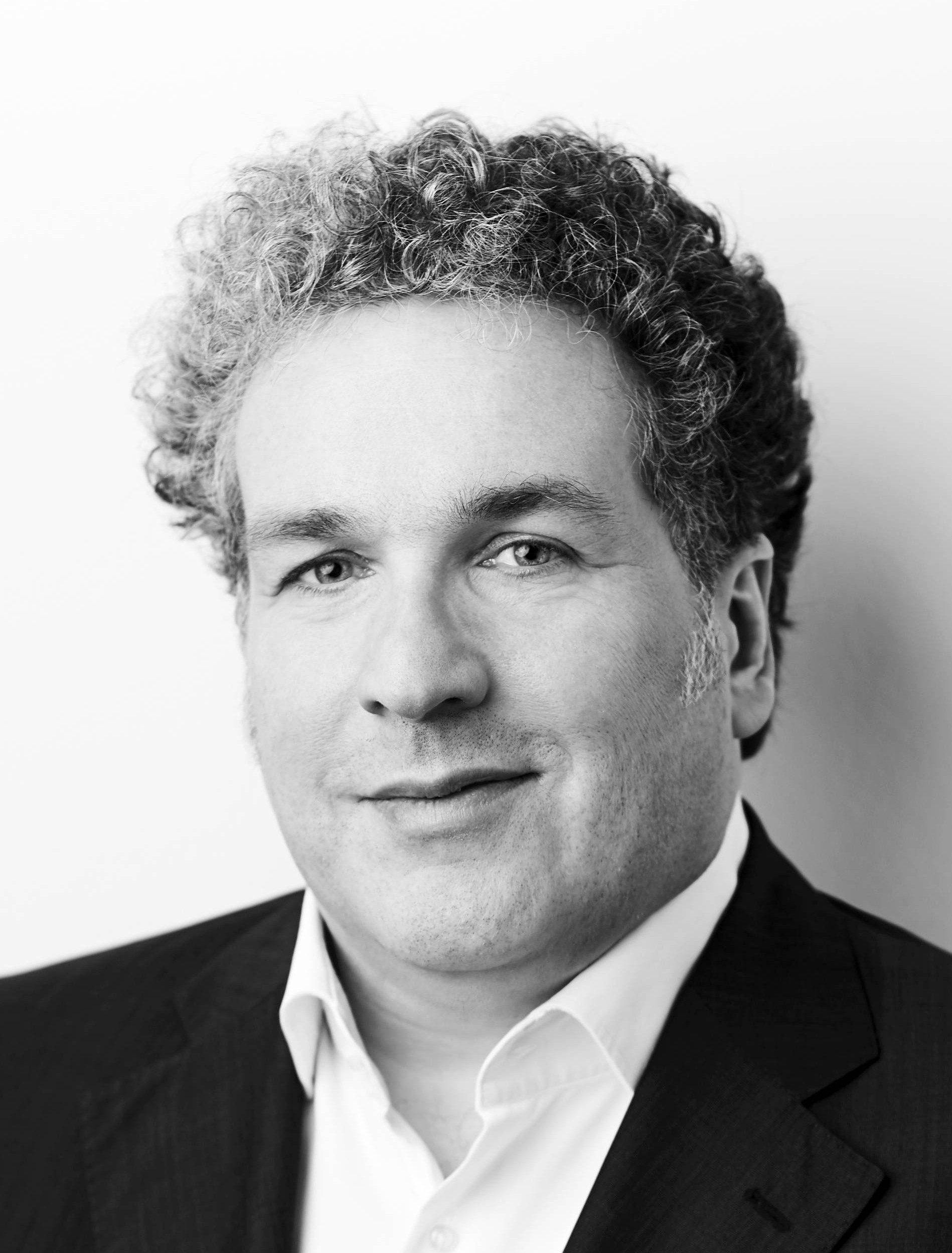}}]
{Oliver Brock}
is the Alexander von Humboldt Professor of Robotics in the School of Electrical Engineering and Computer Science at the Technische Universit{\"a}t Berlin in Germany. He received his Diploma in Computer Science in 1993 from the Technische Universit{\"a}t Berlin and his Master's and Ph.D. in Computer Science from Stanford University in 1994 and 2000, respectively. He also held post-doctoral positions at Rice University and Stanford University. Starting in 2002, he was an Assistant Professor and Associate Professor in the Department of Computer Science at the University of Massachusetts Amherst, before to moving back to the Technische Universit{\"a}t Berlin in 2009. The research of Brock's lab, the Robotics and Biology Laboratory, focuses on autonomous mobile manipulation, interactive perception, grasping, manipulation, soft hands, interactive learning, motion generation, and the application of algorithms and concepts from robotics to computational problems in structural molecular biology. He is also the president of the 
Robotics: Science and Systems foundation.
\end{IEEEbiography}
\begin{IEEEbiography}[{\includegraphics[width=1in,height=1.25in,keepaspectratio]{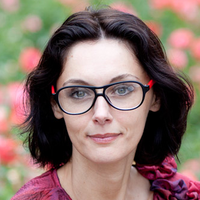}}]
  {Danica Kragic} is a Professor at the School of Computer Science and
  Communication at the Royal Institute of Technology, KTH. She
  received MSc in Mechanical Engineering from the Technical University
  of Rijeka, Croatia in 1995 and PhD in Computer Science from KTH in
  2001. She has been a visiting researcher at Columbia University,
  Johns Hopkins University and INRIA Rennes. She is the Director of
  the Centre for Autonomous Systems. Danica received the 2007 IEEE
  Robotics and Automation Society Early Academic Career Award. She is
  a member of the Royal Swedish Academy of Sciences and Young Academy
  of Sweden. She holds a Honorary Doctorate from the Lappeenranta
  University of Technology. She chaired IEEE RAS Technical Committee
  on Computer and Robot Vision and served as an IEEE RAS AdCom
  member. Her research is in the area of robotics, computer vision and
  machine learning. In 2012, she received an ERC Starting Grant. 
\end{IEEEbiography}
\begin{IEEEbiography}[{\includegraphics[width=1in,height=1.25in,keepaspectratio]{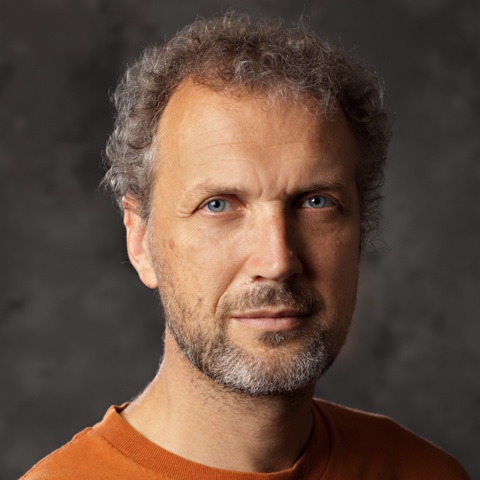}}]
{Stefan Schaal}
is Professor of Computer Science, Neuroscience, and Biomedical Engineering at the University of Southern California, and a Founding Director of the Max Planck Institute for Intelligent Systems in Tuebingen, Germany. In the past, he held positions as a postdoctoral fellow at the Department of Brain and Cognitive Sciences and the Artificial Intelligence Laboratory at MIT, as an Invited Researcher at the ATR Human Information Processing Research Laboratories in Japan, as an Adjunct Assistant Professor at the Georgia Institute of Technology, as an Adjunct Assistant Professor at the Department of Kinesiology of the Pennsylvania State University, and as a group leader of an ERATO Project in Japan.
Dr. Schaal's research interests include topics of statistical and machine learning, neural networks, computational neuroscience, functional brain imaging, nonlinear dynamics, nonlinear control theory, and biomimetic robotics. He applies his research to problems of autonomous systems, artificial and biological motor control, and motor learning, focusing on both theoretical investigations and experiments with human subjects and anthropomorphic robot equipment.
Dr. Schaal is a co-founder of the "IEEE/RAS International Conference and Humanoid Robotics", and a co-founder of "Robotics Science and Systems", a highly selective new conference featuring the best work in robotics every year. Dr. Schaal served as Program Chair at these conferences and he was the Program Chair of "Simulated and Adaptive Behavior" (SAB 2004) and the "IEEE/RAS International Conference on Robotics and Automation" (ICRA 2008), the largest robotics conference in the world. Dr. Schaal is has also been an Area Chair at "Neural Information Processing Systems" (NIPS) and served as Program Committee Member of the "International Conference on Machine Learning" (ICML). Dr. Schaal serves on the editorial board of the journals "Neural Networks", "International Journal of Humanoid Robotics", and "Frontiers in Neurorobotics". Dr. Schaal is a Fellow of the IEEE and a member of the German National Academic Foundation (Studienstiftung des Deutschen Volkes), the Max-Planck-Society, the Alexander von Humboldt 
Foundation, the Society For Neuroscience, the Society for Neural Control of Movement, and AAAS.
\end{IEEEbiography}
\begin{IEEEbiography}[{\includegraphics[width=1in,height=1.25in,keepaspectratio]{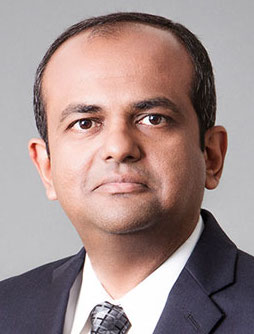}}]
  {Gaurav S. Sukhatme}
  is the Fletcher Jones Professor of Computer
Science (joint appointment in Electrical Engineering) at the
University of Southern California (USC) and Executive Vice Dean of the
USC Viterbi School of Engineering. He received his undergraduate
education at IIT Bombay in Computer Science and Engineering, and
M.S. and Ph.D. degrees in Computer Science from USC. He is the
co-director of the USC Robotics Research Laboratory and the director
of the USC Robotic Embedded Systems Laboratory which he founded in
2000. His research interests are in perception, planning, and robot
networks with applications to environmental monitoring. He has
published extensively in these and related areas. Sukhatme has served
as PI on numerous NSF, DARPA and NASA grants. He was a Co-PI on the
Center for Embedded Networked Sensing (CENS), an NSF Science and
Technology Center. He is a fellow of the IEEE and a recipient of the
NSF CAREER award and the Okawa foundation research award. He is one of
the founders of the Robotics: Science and Systems conference. He was
program chair of the 2008 IEEE International Conference on Robotics
and Automation and the 2011 IEEE/RSJ International Conference on
Intelligent Robots and Systems. He is the Editor-in-Chief of
Autonomous Robots and has served as Associate Editor of the IEEE
Transactions on Robotics and Automation, the IEEE Transactions on
Mobile Computing, and on the editorial board of IEEE Pervasive
Computing. 
\end{IEEEbiography}






\end{document}